# Title

Toward Federated Large Language Models in Medicine: A Parameter-Efficient Framework for Privacy-Preserving, Multi-Institutional Adaptation

## Author list

Anran Li[1], Yuanyuan Chen[2], Wenjun Long[2], Yu Yin[3], Yan Hu[4], Hyunjae Kim[1], Weipeng Zhou[1], Yujia Zhou[1], Hongyi Peng[2], Yang Ren[1], Xuguang Ai[1], Zhenyue Qin[1], Ming Hu[5], Xiaoxiao Li[6], Han Yu[2], Yih-Chung Tham[7], Lucila Ohno-Machado[1], Hua Xu[1], Qingyu Chen[1,*]

**Corresponding author**:

[*]Qingyu Chen, qingyu.chen@yale.edu

## Affiliations

1. Department of Biomedical Informatics and Data Science, School of Medicine, Yale University, New Haven, USA
2. College of Computing and Data Science, Nanyang Technological University, Singapore
3. Department of Earth Science and Engineering, Imperial College London, London, United Kingdom
4. McWilliams School of Biomedical Informatics, University of Texas Health Science at Houston, Houston, USA
5. School of Computing and Information Systems, Singapore Management University, Singapore
6. School of Electrical and Computer Engineering, University of British Columbia, Canada
7. Department of Ophthalmology, Yong Loo Lin School of Medicine, National University of Singapore, Singapore

## 1. Abstract

Large language models (LLMs) are increasingly adapted for medical applications, but most are trained using data from a single institution because privacy and governance constraints prevent multi-institutional data sharing. As a result, these models often generalize poorly across heterogeneous healthcare systems. We address this gap by introducing Fed-MedLoRA and Fed-MedLoRA+, a parameter-efficient federated framework for collaborative LLM adaptation across healthcare institutions. Fed-MedLoRA transmits only low-rank adapters rather than full model weights, reducing communication overhead. We also evaluate a privacy-preserving variant that applies Gaussian perturbation to transmitted adapter updates. Fed-MedLoRA+ further incorporates adaptive aggregation to better address cross-site heterogeneity in patient populations, annotation practices, and disease distributions.

We evaluate the framework on clinical information extraction across five independent patient cohorts totaling 42,198 entities and 41,570 relations, and compare it with zero-shot and fine-tuned LLMs, domain-specific BERT models, and federated baselines. Across all settings, the proposed methods consistently improve extraction performance and generalize better to heterogeneous cohorts. In a real-world case study using clinical notes from the Yale New Haven Health System, the framework demonstrates strong performance under low-resource new-site deployment. These results suggest that federated, parameter-efficient LLM adaptation is feasible, scalable, and effective for multi-institutional clinical deployment.

## 2. Introduction

LLMs are rapidly transforming the landscape of medicine, with extensive studies demonstrating their potential across a broad spectrum of applications, including outpatient management [1,2], disease diagnosis [3,4], and clinical summarization [5,6]. To support clinical adoption, growing efforts have focused on adapting general-domain LLMs through continued pretraining or post training on clinical data to improve accuracy and safety in downstream medical tasks [7–9].

However, in practice, many LLMs are trained or adapted using clinical data from a single institution, primarily due to privacy and governance constraints that make sharing patient-level data across healthcare systems challenging [10–13]. Yet clinical data vary substantially across institutions in patient demographics, disease prevalence, care pathways, documentation styles, and clinical practices, leading to domain shift and degraded model generalization [14–16]. Recent studies have highlighted that this represents a significant—and often underestimated—limitation in the development and deployment of LLMs in medicine [17,18]. For example, the accuracy of a locally fine-tuned LLM for hospital admission prediction dropped by more than 20% when evaluated on external patient populations [18].

Federated learning (FL) offers a privacy-preserving paradigm in which institutions collaboratively train models without sharing raw patient data, thereby directly addressing regulatory and governance constraints in healthcare [19]. In a standard FL workflow, models are trained locally at each site, model parameters are transmitted to a central server for aggregation, and the updated global model is redistributed for further refinement [20,21]. However, adapting FL to medical LLMs remains fundamentally challenging [22–24]. First, conventional FL methods such as FedAvg require transmitting full model parameters at each communication round, which is infeasible for multi-billion–parameter LLMs given the constrained computational and networking resources available in most healthcare environments [25–27]. Second, in real-world clinical settings, data are often highly heterogeneous across institutions, with differences in patient populations, disease prevalence, annotation availability, and documentation practices. Under these conditions, FedAvg-style aggregation [28,29] of heterogeneous local model parameters can lead to client drift and uneven site-wise performance, because local objectives and optimization directions are inconsistent across sites [25,30].

To date, although recent perspectives have highlighted the promise of federated LLMs in medicine, concrete methodological solutions remain limited [31,32]. Existing medical FL studies have largely focused on smaller architectures such as BERT [33] or traditional classifiers such as logistic regressions [34], which do not face the same scalability constraints as LLMs. Beyond the medical domain, communication-efficient distributed training methods, such as DiLoCo [35], have demonstrated that infrequent synchronization can significantly reduce communication overhead during large-model optimization. However, these approaches are primarily designed for distributed pretraining across loosely connected compute clusters and do not address the privacy and heterogeneity constraints in cross-institutional medical learning. Similarly, recent LoRA-based federated frameworks (e.g., FDLoRA [36]) have focused on personalized training in general domains. In addition, federated foundation model efforts in medical imaging (e.g., FFMed [32]) have demonstrated the importance of heterogeneity-aware optimization, but are largely limited to vision modalities. Other systems, such as Ollama-based medical AI frameworks [37], emphasize local inference and deployment integration of pretrained models rather than cross-silo collaborative training. Therefore, parameter-efficient frameworks for federated training of LLMs in medicine remain underdeveloped.

To address these challenges, we introduce Fed-MedLoRA and Fed-MedLoRA+, a parameter-efficient federated learning framework for training LLMs in medicine. Fed-MedLoRA, extending federated optimization to LoRA adapter parameters, enables scalable federated training by transmitting only low-rank adapter parameters, substantially reducing communication and computation overhead while preserving model capacity. We also apply Gaussian perturbation to transmitted adapter updates as a privacy-enhancing mechanism. Building on this foundation, Fed-MedLoRA+ incorporates an adaptive, data-aware aggregation strategy that explicitly accounts for cross-site heterogeneity, thereby improving global model convergence under realistic multi-institutional conditions. More broadly, our contribution is a practical and scalable adaptation of federated learning principles to the constraints of medical LLM deployment.
We evaluate the proposed framework using clinical information extraction (IE) as a representative downstream application. Clinical IE involves identifying and normalizing key medical entities, capturing relations among them, and mapping extracted information to standardized knowledge representations across healthcare systems [38,39]. Despite its importance for critical downstream applications—automated cohort identification [23], adverse event monitoring [24], and clinical decision support [40]—prior studies have shown that current LLMs exhibit suboptimal performance on clinical IE [11,41–43]. We systematically evaluated the proposed framework for accuracy and practical feasibility. Accuracy was assessed across five independent patient cohorts totaling 42,198 entities and 41,570 relations on two core IE tasks—named entity recognition (NER) and relation extraction (RE)—under three settings: (1) in-domain training and testing to assess within-domain performance; (2) independent benchmark validation to evaluate generalization to external patient cohorts; and (3) low-resource new-site adaptation using a case study based on real-world clinical notes from the Yale New Haven Health System to simulate model deployment at new institutions with limited annotations. Both Fed-MedLoRA and Fed-MedLoRA+ were implemented using two representative open-weight LLMs—Llama-3.1-8B-Instruct and DeepSeek-R1-Distill-Llama-8B—as backbones and were compared head-to-head against multiple baselines: zero-shot and single-site fine-tuned LLMs (Llama-3.1-8B-Instruct, DeepSeek-R1-Distill-Llama-8B, and zero-shot GPT-4o), fine-tuned domain-specific BERT models, and recent general-domain FL algorithms for LLMs. For practical feasibility, we further assessed (1) robustness to heterogeneous and incomplete task annotations across sites (e.g., one site with both NER and RE vs. another with only NER), (2) computational efficiency during training and inference, (3) scalability as the number of participating sites increased, and (4) the trade-off between utility and privacy robustness.

Across all evaluation settings, Fed-MedLoRA and Fed-MedLoRA+ consistently outperformed the baselines for both NER and RE tasks. In the in-domain benchmarks, they improved zero-shot LLM performance by up to 65% absolute F1, achieved ~25% higher F1 than single-site fine-tuned LLMs, and outperformed fine-tuned domain-specific BERT models by up to 40%. In external validation across independent cohorts, they maintained strong generalization and had 10-70% higher F1 than the baselines. In the low-resource new-site adaptation study, Fed-MedLoRA+ had the best performance, suggesting that a new clinical site could benefit from federated LLMs to bootstrap local model development with limited annotations. Beyond accuracy, the framework proved feasible and scalable in practical settings. Transmitting only low-rank adapters reduced communication cost by 98.5% relative to full-model fine-tuning. The framework scaled efficiently to ten participating sites, and remained robust when sites contributed uneven or incomplete task annotations, reflecting realistic multi-institutional conditions.

Overall, the results demonstrate that federated, parameter-efficient adaptation could make medical LLMs more practical, scalable, and robust under privacy, heterogeneity, and resource constraints. More broadly, this work establishes a concrete framework for federated LLM development in medicine and provides empirical guidance for future multi-institutional deployment. To promote transparency, reproducibility, and community collaboration, all implementation code and documentation are publicly available at [https://github.com/Yale-BIDS-Chen-Lab/FL_LLM_Med](https://github.com/Yale-BIDS-Chen-Lab/FL_LLM_Med).

# 3. Results

In this section, we systematically evaluated both accuracy and practical feasibility of Fed-MedLoRA and Fed-MedLoRA+.

### 3.1 Evaluation overview

Table 1 summarizes the five cohorts used in this study. Collectively, those five cohorts consist of 42,198 entities and 41,570 relations covering four main entity categories and 16 relation types.

We used four existing benchmarks manually annotated from different institutions: MIMIC-III (Medical Information Mart for Intensive Care) [44], MTSamples (Transcribed Medical Transcription Sample Reports and Examples) [45], UTP (UT Physicians notes) [16], and Partners Healthcare and Beth Israel Deaconess Medical Center notes by Informatics for Integrating Biology & the Bedside (I2B2) [46]. These datasets have been widely adopted for clinical IE method development and evaluation [16,44–46]. They reflect real-world challenges, including differences in scale (e.g., MIMIC-III notes are significantly longer than those from other datasets), entity distributions (e.g., MIMIC-III has nearly twice as many drug entities as UTP due to different patient cohorts), and missing data (e.g., MTSamples lacks reference ranges).

In addition, we manually annotated 242 problem-related entities, 27 treatment-related entities, 175 test-related entities, and 44 drug entities from 291 sentences drawn from four de-identified patient records within the Yale New Haven Health (YNHH) system for the case study. The annotation scale was intentionally limited to reflect a realistic starting point for a new clinical site compared with established benchmarks. Please see Supplementary Material S3.1 for more information on dataset preprocessing and annotation.

The system overview is illustrated in Figure 1. As shown in Figure 1B, we systematically evaluated the accuracy of Fed-MedLoRA and Fed-MedLoRA+ under three settings. First, we conducted standard training and testing. Models were trained on the training sets and evaluated on the corresponding test sets to assess in-domain performance. We conducted both two-site (FL across MIMIC-III and MTSamples) and three-site (MIMIC-III, MTSamples, and UTP) experiments under this setting. Second, we evaluated generalization on independent benchmarks. To evaluate performance on unseen distributions, models trained under the standard training and testing setting were directly tested on external datasets. In the two-site experiment (training on MIMIC-III and MTSamples), we evaluated performance on UTP and I2B2. In the three-site experiment (training on MIMIC-III, MTSamples, and UTP), we evaluated on I2B2. Third, we conducted a low-resource case study. To simulate a practical scenario where a new clinical site performs clinical IE locally, we conducted a case study using the YNHH cohort. Unlike the standard benchmarks in the previous settings, a new site typically has only minimal annotated data available for initialization and needs to adapt existing models for local development.
As shown in Figure 1C, we reported precision, recall, and F1-score, the standard metrics for NER and RE [25,47]. Following prior work [16,25], we used two evaluation schemes. (1) Strict match: for NER, a true positive requires exact span and entity type match; for RE, this further requires the correct entity pair and relation type. (2) Lenient match: for NER, a true positive requires overlapping spans and correct entity type; for RE, relations are considered correct when constituent entities overlap with ground-truth spans and the relation type matches. We also conducted a manual error analysis to complement the automatic metrics and categorize the errors [48,49].

Beyond accuracy, we further evaluated the practical feasibility of Fed-MedLoRA and Fed-MedLoRA+ across three aspects. Figure 1D shows the details. First, robustness to uneven task annotations across sites. In practice, individual sites may lack complete annotations for all tasks. We conducted both two-site and three-site experiments to examine this scenario. In the two-site experiment, Site A (MIMIC-III) included annotations for both NER and RE, whereas Site

B (MTSamples) had only NER annotations. In the three-site experiment, Site C (UTP) contained only NER annotations, while Sites A and B provided both NER and RE. Second, communication and computational efficiency. We quantified memory consumption and peak GPU utilization during both training and inference to assess efficiency. Third, scalability. We conducted simulations with up to ten participating sites to evaluate the scalability under increasing federation sizes.

We compared Fed-MedLoRA and Fed-MedLoRA+ against three categories of baselines: (1) BERT-based models, (2) LLM-based models, and (3) recent general-domain FL algorithms for LLMs. For BERT-based baselines, we used Bio_ClinicalBERT [50,51], initialized from BioBERT (pretrained on biomedical literature) [52] and further pretrained on clinical notes. Bio_ClinicalBERT has achieved completive performance on multiple clinical IE benchmarks and is widely adopted as a strong baseline [51]. We reported its single-site fine-tuning performance. Single-site training reflects a common scenario in clinical practice, where models are trained on data from a single site without federated learning support. For instance, for three datasets from different sites, we trained a model on one site's training set and evaluated it on all test sets, repeating for each site. The reported results are averaged across sites.

For LLM-based baselines, we performed head-to-head comparisons using the same backbone models with and without our proposed federated learning methods. Fed-MedLoRA and Fed-MedLoRA+ are backbone-agnostic; we selected two representative open-weight LLMs Llama-3.1-8B-Instruct [53] and DeepSeek-R1-Distill-Llama-8B [54] for evaluation. We reported both zero-shot (baseline) and single-site (average across sites, baseline) fine-tuned performance. Additionally, we included GPT-4o as a widely used proprietary LLM representative and reported its zero-shot performance for comparison. To complement this evaluation, we included FedSA-LoRA [55] as a representative general-domain FL baseline for LLMs. This method reduces communication by transmitting only the low rank $A$ to the server for aggregation, under the assumption that the $A$ matrices capture the global knowledge. In addition, we also included centralized training for both BERT-based and LLM-based models as a reference. Centralized training pools data from all sites to train a single model, which is then evaluated on the same test sets. This configuration provides an empirical upper bound, representing an ideal but impractical setting in healthcare due to privacy and governance constraints. The implementation and hyperparameter details are provided in Section 5 Methods.

For both the training/testing and independent benchmark settings, we performed bootstrapping with a sample size of 200 randomly selected instances per dataset and repeated 30 times We reported results with 95% confidence intervals to quantify uncertainty. Since model performances were evaluated on the same test sets and under the same bootstrap resamples, statistical comparisons are paired rather than independent. Therefore, we used a two-tailed Wilcoxon signed-rank test to assess statistical significance between paired model performance estimates on the same test sets. Both procedures followed established practices [56,57].

### 3.2 Results on accuracy

Table 2 summarizes model performance in the two-site experiment, where federated learning was conducted across MIMIC-III and MTSamples and evaluated on their respective test sets.

Table 3 presents corresponding results for the three-site experiment involving MIMIC-III, MTSamples, and UTP under the same training/testing setting. More results including individual entity accuracy and Bootstrapping and statistical analysis are illustrated in Supplementary S2.1.

The proposed Fed-MedLoRA and Fed-MedLoRA+ consistently outperformed LLM-based baselines (both zero-shot and single-site fine-tuned) using the same backbone models (Llama-3.1-8B-Instruct, DeepSeek-R1-Distill-Llama-8B and GPT-4o) in head-to-head comparisons. Compared to zero-shot performance, Fed-MedLoRA and Fed-MedLoRA+ achieve over 50% absolute improvement on both NER and RE test sets. For example, in the two-site Llama-3.1-8B-Instruct experiment on MIMIC-III, strict matching NER F1 score improves from 0.345 (zero-shot) to 0.850 (Fed-MedLoRA+, $p<0.0001$), while RE strict F1 score improves from 0.203 (zero-shot) to 0.860 (Fed-MedLoRA+, $p<0.0001$). Similar trends are observed in the three-site experiment. Although GPT-4o achieved the highest zero-shot NER performance among all evaluated LLMs on the MTSamples dataset (strict and lenient F1 scores of 0.602 and 0.834, respectively), its zero-shot performance was still significantly lower than that of federated fine-tuning models. These results demonstrate that federated instruction tuning can significantly improve performance on NER and RE tasks.

In addition, compared to the LLM-based single-site baseline—which reflects the most common clinical scenario where models are trained using data from a single institution—Fed-MedLoRA+ consistently outperform across all test sets in both two-site and three-site experiments. The largest absolute gains are observed in RE, with up to ~24.2% absolute improvement. For example, in the two-site experiment on MIMIC-III, the single-site baseline with Llama-3.1-8B-Instruct achieves 0.648 strict F1 score, compared to 0.860 for Fed-MedLoRA+ ($p<0.0001$). Similarly, with DeepSeek-R1-Distill-Llama-8B, the single-site baseline achieves 0.589 versus 0.831 ($p<0.0001$) for Fed-MedLoRA+. This improvement may relate to the fact that RE is inherently more challenging than NER, and leveraging data from multiple sites provides greater benefit. We elaborate on this in the section Comparisons between tasks. In terms of backbone models, Llama-3.1-8B-Instruct and DeepSeek-R1-Distill-Llama-8B demonstrate similar performance, with Llama-3.1-8B-Instruct slightly outperforming in the three-site experiments.

Centralized learning pools data from all sites for training which is impractical for medical applications. We use it as a reference of estimated upper bound. As shown in Tables 2 and 3, Fed-MedLoRA and Fed-MedLoRA+ achieve performance competitive with centralized training. Notably, Fed-MedLoRA+ closely approach centralized learning across most datasets under both two-site and three-site experiments. For example, in the two-site experiment for NER, Fed-MedLoRA+ achieves strict F1 scores of 0.850 vs. 0.856 on MIMIC-III for centralized learning, and 0.866 vs. 0.870 ($p=0.081$) on MTSamples. In the three-site experiment, the gaps are slightly larger, e.g., 0.803 vs. 0.840 ($p=0.042$), 0.862 vs. 0.867 ($p=0.121$), and 0.898 vs. 0.909 ($p=0.076$), but Fed-MedLoRA+ remains the closest to the centralized upper bound.

Fed-MedLoRA and Fed-MedLoRA+ consistently outperform BERT-based models, with gains of up to 9.7% absolute F1 score on NER (e.g., 0.850 for Fed-MedLoRA+ with Llama-3.1-8B-Instruct vs. 0.753 for single-site Bio_ClinicalBERT, $p<0.001$) and up to 42.6% absolute F1 score on RE (e.g., 0.860 for Fed-MedLoRA+ with Llama-3.1-8B-Instruct vs. 0.434 for Bio_ClinicalBERT, $p<0.0001$). Interestingly, when comparing BERT-based and LLM-based single-site baselines,

LLM-based models outperform BERT across nearly all test sets, except for a marginally lower performance on MIMIC-III in the three-site experiment (0.775 vs. 0.782, p=0.069). We conducted a more detailed comparative analysis between BERT-based and LLM-based models, combining results from both in-domain and independent benchmark settings, as presented below.

As mentioned above, we further compared the proposed methods with FedSA-LoRA [58], a recently introduced FL algorithm for LLMs in the general domain, used here as an additional baseline. Table 4 presents head-to-head comparisons from the three-site experiment using the Llama-3.1-8B-Instruct model. Across all evaluation datasets, Fed-MedLoRA and Fed-MedLoRA+ consistently outperformed FedSA-LoRA on both NER and RE tasks. For example, FedSA-LoRA achieved a strict F1 score of 0.298 on NER UTP, compared to 0.898 for Fed-MedLoRA+. Similarly, FedSA-LoRA obtained a strict F1 score of 0.181 for RE, compared to 0.894 for Fed-MedLoRA+. The two-site experiment results showed similar trends (see Supplementary S2.1). We anticipate two possible reasons for these improvements. First, the proposed Fed-MedLoRA and Fed-MedLoRA+ exchange task-aware, parameter-efficient adapters that preserve task-specific information across sites, whereas FedSA-LoRA transmits only a single low-rank matrix, potentially losing important local adaptation signals. Second, FedSA-LoRA was not originally designed for clinical applications, whereas Fed-MedLoRA+ incorporates influence-aware aggregation to enhance robustness against cross-site heterogeneity.

Across both two-site and three-site experiments, Fed-MedLoRA+ generally outperforms Fed-MedLoRA. In the two-site experiment, Fed-MedLoRA+ achieves higher scores on all four test sets (NER and RE on both datasets). In the three-site experiment, Fed-MedLoRA+ outperforms Fed-MedLoRA on four out of six test sets, with only marginal differences in the remaining two sets (e.g., 0.803 vs. 0.809 on MIMIC-III, p=0.892, and 0.913 vs. 0.916 on UTP, p=0.104). Collectively, across the two- and three-site experiments, both Fed-MedLoRA and Fed-MedLoRA+ consistently outperforms the baselines except for centralized training methods. An important observation is that adding more sites does not always guarantee better performance on every site. For instance, both Fed-MedLoRA and Fed-MedLoRA+ achieve higher test scores on MIMIC-III under the two-site experiment than under the three-site experiment. A similar pattern is observed for centralized learning, which serves as the estimated upper bound: the exact F1 score on MIMIC-III decreases from 0.856 in the two-site experiment to 0.840 in the three-site experiment. This suggests that while incorporating multiple sites generally increases robustness relative to single-site learning, it does not guarantee uniform gains on every test set. This aligns with prior findings, which highlight challenges such as data heterogeneity, domain shifts, and label distribution differences. One plausible explanation is that the third site introduces increased data heterogeneity: entity distribution differences and annotation noises (see manual error analysis results in Table 7) may bias aggregate updates and create trade-offs between sites.

Another observation is that results on RE test sets are generally lower than those on NER. Indeed, NER focuses on single-entity recognition, while RE requires identifying entity pairs and the semantic relation between them, which is inherently more challenging [59]. Fed-MedLoRA and Fed-MedLoRA+ demonstrate greater robustness on RE compared to baselines. For example, in the three-site experiment, the single-site Llama-3.1-8B-Instruct baseline achieves 0.775 on NER but drops to 0.656 on RE. In contrast, Fed-MedLoRA+ achieves an NER score of 0.803 and a RE score of 0.771. A similar pattern is observed with DeepSeek-R1-Distill-Llama-8B: the single-site baseline performance drops from 0.778 on NER to 0.605 on RE, whereas Fed-MedLoRA+ achieves 0.802 on NER and 0.762 on RE. Thus, aggregating distributed and diverse training samples via federated LoRA substantially reduces the gap of RE relative to single-site training.

To assess whether the observed performance gains are attributable to output formats rather than extraction capability, we conduct the two-site experiments using a JSON output template instead of the original HTML template. As shown in Table 5, results under the JSON template are highly consistent with those obtained with the original HTML template across sites and training strategies. The differences between the two formats are small, typically within 1% F1 score. In particular, Fed-MedLoRA and Fed-MedLoRA+ outperform the zero-shot, single-site and Bio_ClinicalBERT baselines under both output formats, and Fed-MedLoRA+ remains the best non-centralized method in terms of both strict and lenient F1 scores. The results for RE are much similar to those for NER (see Supplementary S2.1 and Table S3).

Table 6 summarizes the results of the two-site experiment (federated learning across MIMIC-III and MTSamples, evaluated on UTP and I2B2) and the three-site experiment (federated learning across MIMIC-III, MTSamples, and UTP, evaluated on I2B2) under the independent benchmark setting. Supplementary S2.1 provides additional results on individual entity accuracy.

Similar to the standard training/testing setting, both Fed-MedLoRA and Fed-MedLoRA+ consistently outperform both zero-shot and single-site fine-tuning baselines on independent benchmarks using the same LLM backbones. Note that zero-shot baselines remain unchanged across settings since no fine-tuning is applied. For NER, Fed-MedLoRA+ achieves ~58.4% absolute gain (e.g., exact F1 score of 0.256 for zero-shot vs. 0.840 for Fed-MedLoRA+ on UTP using Llama-3.1-8B-Instruct) and ~70.2% absolute gain on RE (e.g., 0.091 vs. 0.793 in the same setting). Similar gains are observed with DeepSeek-R1-Distill-Llama-8B and GPT-4o as backbones. When compared to single-site fine-tuning, the federated approach also has consistent improvements. For instance, in the three-site experiment on I2B2 with Llama-3.1-8B-Instruct, Fed-MedLoRA+ achieves up to a 9.6% gain on NER (strict F1 score: 0.852 for Fed-MedLoRA+ vs. 0.756 for single-site fine-tuning) and a 5.4% gain on RE (strict F1 score: 0.713 for Fed-MedLoRA+ vs. 0.659 for single-site fine-tuning).

The performance gap between federated approaches and BERT-based baselines was more significant under independent benchmarks compared to the train/test testing. For NER, Fed-MedLoRA+ achieved ~15% higher exact F1 than single-site fine-tuned Bio_ClinicalBERT (UTP: 0.840 vs. 0.678; I2B2: 0.860 vs. 0.747) in the two-site experiment, and over 10% higher in the three-site experiment (I2B2: 0.852 vs. 0.752). For RE, Fed-MedLoRA+ achieved improvements of ~70% on UTP (0.793 vs. 0.270) and ~50% on I2B2 (0.740 vs. 0.302) in the two-site experiment, and a similar ~50% gain on I2B2 in the three-site experiment (0.713 vs. 0.219). Centralized learning results also revealed consistent patterns: centralized LLMs substantially outperformed centralized BERT models on independent benchmarks. Collectively, when comparing LLMs vs. BERT under both training/testing and independent benchmarks, the performance gap was much larger on external datasets. For instance, under the training/testing setting, centralized

LLMs achieved only ~4% higher F1 than BERT on NER (e.g., 0.865 vs. 0.807 on MIMIC-III; 0.870 vs. 0.827 in the two-site experiment). In contrast, under external benchmarks, centralized LLMs had ~16% higher F1 on UTP (0.852 vs. 0.692) and ~9.5% higher on I2B2 (0.863 vs. 0.768). These results suggest that while BERT models remain competitive when training and testing distributions are similar, they generalize poorly to external datasets. This observation is consistent with prior reports in the literature[16,60,61].

Additional observations are consistent with the training/testing results. First, the proposed federated approaches consistently achieved performance closest to centralized learning as the estimated upper bound (e.g., ~2% difference between Fed-MedLoRA+ and centralized learning on both UTP and I2B2 for NER and RE in the two-site experiment). Second, Fed-MedLoRA+ generally outperformed Fed-MedLoRA (e.g., higher scores in all 8 metrics—strict and lenient F1, NER and RE, on UTP and I2B2 in the two-site experiment). Third, Llama-3.1-8B-Instruct and DeepSeek-R1-Distill-Llama-8B showed similar performance, with Llama-3.1-8B-Instruct slightly better for both NER and RE. In contrast, we also observed mixed effects of increasing the number of sites. Under the training/testing setting, performance in the two-site experiments was generally higher than in the three-site experiments. When applied to external benchmarks, some performance drop persisted but was less consistent. For example, with Llama-3.1-8B-Instruct on I2B2 NER, the single-site fine-tuned baseline dropped from 0.801 (two-site) to 0.756 (three-site), while Fed-MedLoRA remained robust (0.860 vs. 0.852).

As described above, we conducted a case study using the newly annotated YNHH cohort to simulate a practical scenario in which a new site performs local clinical IE. In contrast to established benchmarks, such a site typically has only minimal annotated data available for initialization and must rely on adapting existing models rather than fine-tuning locally. Realistic options therefore include using publicly available LLMs in a zero-shot manner or leveraging federated models trained on data from other institutions without sharing sensitive information.

We directly applied Fed-MedLoRA and Fed-MedLoRA+ models trained in the two-site and three-site experiments, using Llama-3.1-8B-Instruct as the backbone. For comparison, we also evaluated the zero-shot Llama-3.1-8B-Instruct baseline, along with centralized Llama-3.1-8B-Instruct and Bio_ClinicalBERT, which serve as empirical upper-bound references. Evaluation metrics were consistent with those described above. Table 7 summarizes the results. Overall, both Fed-MedLoRA and Fed-MedLoRA+ consistently outperformed all baselines on the new site annotations. For instance, directly applying Fed-MedLoRA+ from the three-site experiment achieved the highest performance, with a strict F1-score of 0.730 compared to 0.397 for the zero-shot baseline ($p<0.0001$), and a lenient F1-score of 0.854 compared to 0.524 for the zero-shot baseline ($p<0.0001$). While Fed-MedLoRA also demonstrated robust performance, Fed-MedLoRA+ consistently had higher scores under both strict and lenient evaluation criteria. Moreover, Fed-MedLoRA+ achieved performance comparable to the upper-bound centralized Llama-3.1-8B-Instruct model (e.g., 0.720 vs. 0.732 and 0.730 vs. 0.733 in the two-site and three-site experiments, $p=0.0691$, $p=0.0874$). Notably, both federated models also outperformed the centralized Bio_ClinicalBERT baseline, consistent with earlier observations on the independent benchmark setting. Results of individual entity performance are shown in Supplementary S2.2.

We further conducted a manual error analysis to characterize the common errors in the NER predictions of Fed-MedLoRA using the Llama-3.1-8B-Instruct backbone. From the four test corpora (MIMIC-III, MTSamples, UTP, and I2B2), we randomly sampled 200 error instances for qualitative review and categorized consistent with prior studies [16]. Table 8 summarizes the identified error categories, their frequencies, and representative examples. Because a single instance may contain multiple error types, the categories are not mutually exclusive and the reported frequencies do not sum to 100%

In the sampled error set, boundary/span errors were the most common failure mode (90/200, 45.0%), followed by false negatives (72/200, 36.0%), type confusion (34/200, 17.0%), false positives (18/200, 9.0%), boundary and type errors (16/200, 8.0%), annotation errors (9/200, 4.5%), and merged/split entities (5/200, 2.5%). Overall, boundary errors and false negatives were the most frequent failure modes, followed by type confusion and false positives. Boundary errors typically occurred when the model predicted spans that overlapped with, but did not exactly match, the ground-truth spans—resulting in penalties under the strict F1 metric. False negatives frequently reflected omissions of shorter or nested entities.

### 3.3 Results on practical feasibility

Beyond accuracy, we further evaluated model robustness under a practical scenario in which different sites provided annotations for different tasks. Table 9 summarizes results from both two-site and three-site experiments. (In the two-site setting, Site A (MIMIC-III) included annotations for both Named Entity Recognition (NER) and Relation Extraction (RE), whereas Site B (MTSamples) included only NER annotations. In the three-site setting, Site C (UTP) contained only NER annotations, while Sites A and B provided both NER and RE.) Importantly, both single-site fine-tuned BERT and LLM baselines are not applicable in scenarios with uneven task annotations. We compared Fed-MedLoRA and Fed-MedLoRA+ trained under these uneven annotation conditions directly with the models trained using complete task annotations.

As shown in Table 9, both Fed-MedLoRA and Fed-MedLoRA+ maintained strong performance despite missing task annotations across sites, consistently outperforming single-site fine-tuning baselines that used complete annotations. Compared with their fully annotated counterparts, Fed-MedLoRA+ had only a slight decrease in NER performance (1.2%–1.7% reduction in F1-score across test sets). For RE, the performance drops were somewhat larger but still moderate, with absolute F1-score decreases of 2.8%–5.8%.

In addition, we evaluated the communication overhead and computational resource requirements of Fed-MedLoRA and Fed-MedLoRA+. Specifically, we report per-round communication volumes (number of parameters transmitted and their storage size), peak GPU memory consumptions for training and inference, and total training time (in GPU-hours). To better quantify the trade-off between model accuracy and computational efficiency, we additionally trained Fed-MedLoRA and Fed-MedLoRA+ using Llama-3.2-1B as the backbone, in contrast to the 8B model used in the primary evaluations.

Figure 2A compares the communication overhead of Fed-MedLoRA+ against full-parameter fine-tuning. Recall that both Fed-MedLoRA and Fed-MedLoRA+ update only the low-rank adapters, rather than the full model parameters. For Llama-3.1-8B-Instruct model (8,030,261,248 parameters), full-parameter fine-tuning requires about 29.92 GB per site per round in float 32. This translates to a total transmission of 239 GB for the two-site experiment and 359 GB in the three-site experiment. In contrast, Fed-MedLoRA and Fed-MedLoRA+ update only 41,943,040 parameters, corresponding to a 99.48% reduction in transmission volume. The resulting transmission requirements are 1.25 GB for the two-site experiment and 1.88 GB for the three-site experiment. Similar reductions are observed for other LLM backbones.

Figure 2B reports the peak GPU memory usage of Fed-MedLoRA+ during training. Peak GPU memory consumption exceeds the raw parameter storage because additional memory is required for gradients, activation checkpoints, optimizer states (e.g., momentum and variance estimates). For Fed-MedLoRA+ with Llama-3.1-8B-Instruct, peak training memory was ~14.04 GB, making it feasible to train on a single consumer-grade GPU card such as the NVIDIA RTX 4090 (16 GB)—without requiring high-end datacenter GPUs like the H100 or H200, which may not be readily available in clinical environments. For inference, since gradients and optimizer states are not present, the memory footprint was substantially lower—~6.79 GB, depending on batch size and sequence length. In practical terms, this lower inference footprint means that models trained using Fed-MedLoRA+ can be deployed on widely available GPUs such as the RTX A6000 (12–16 GB) and RTX 3090/4090, which is crucial for clinical sites that lack training infrastructure but can perform local inference [62].

Figure 2C presents the wall-clock training time in GPU-hours. For Fed-MedLoRA+ with Llama-3.1-8B-Instruct, the total training time was 6.89 GPU-hours for the two-site experiment and 6.98 GPU-hours for the three-site experiment. Comparable runtimes were observed for DeepSeek-R1-Distill-Llama-8B. We also provided total training FLOPs comparison in Supplementary S2.2. Table 10 compares the performance of Fed-MedLoRA and Fed-MedLoRA+ when using Llama-3.2-1B versus the larger 8B backbone models. Overall, the smaller 1B backbone had a ~3% decrease in NER accuracy and up to a 7% decrease in RE accuracy. As illustrated in Figure 2, the reduction in accuracy comes with substantial gains in efficiency. The 1B model reduced per-round communication cost (e.g., 344 MB vs. 1.25 GB in the two-site experiment), GPU memory cost during training (2.15 GB vs. 14.04 GB) and inference (1.24 GB vs. 6.83 GB), and total GPU hours (1.82 h vs. 6.89 h). In practical terms, training Fed-MedLoRA and Fed-MedLoRA+ with the 1B model can be performed on widely available GPUs such as the NVIDIA RTX 3060 Ti or RTX 4060 Ti (8 GB), while inference can be executed on standard or institution-managed laptops (e.g., devices equipped with an Apple M3 Pro–class chip or an equivalent x86 GPU with over 18 GB RAM). These results suggest the potential feasibility of FL with smaller LLM backbones in resource-constrained environments, where a modest decrease in accuracy is acceptable.

### 3.6 Assessment of model scalability

Motivated by the above results, we further conducted simulations with up to $k$ = 10 participating sites to assess scalability. For each $k$, the training data from MIMIC-III, MTSamples, and UTP were partitioned into $k$ random shards, with each shard treated as an independent site. Consistent with prior experiments, we compared the strict F1-scores of Fed-MedLoRA+ and single-site fine-tuned LLM baselines using Llama-3.2-1B as the backbone for both NER and RE, with centralized learning serving as the estimated upper-bound reference under both training/testing and independent benchmark settings.

Figure 3 summarizes the results. Overall, Fed-MedLoRA+ maintained performance nearly identical to centralized learning when $k \le 5$ and had only moderate degradation as $k$ increased to 10. The absolute performance drops at $k = 10$ is 3.9% for NER and 7.5% for RE compared to centralized learning. In contrast, the single-site fine-tuned LLM baseline showed sharp degradation as $k$ increased. For instance, on the MIMIC-III dataset, Fed-MedLoRA+ achieved a 21.8% higher F1-score than the single-site baseline for RE at $k$ =10.

We also observed that RE is inherently more challenging and more sensitive to the number of participating sites than NER, which is consistent with our findings across both training/testing and independent benchmark settings, where RE consistently achieved lower scores than NER. In simulations with up to 10 sites, NER micro F1-scores remained above 0.79 across all test sets, whereas RE showed greater variability, with a minimum F1-score of approximately 0.618 on the I2B2 dataset.

### 3.7 Privacy preservation ablation study results

To evaluate the empirical utility-privacy trade-off introduced by Gaussian perturbation on transmitted LoRA parameters, we varied the noise scale σ from 0 to 0.01 using the LLaMA-3.1-8B-Instruct backbone for Fed-MedLoRA under the two-site setting. Fed-MedLoRA+ exhibited a similar trend under increasing noise levels. As shown in Table S4, model performance remained highly stable under slight perturbations. Specifically, when $\sigma$ increased from 0 to 0.001, the strict and lenient F1 scores changed marginally across all test sets. Increasing the noise to $\sigma$=0.005 led to a modest performance drop, but the model still preserved high NER accuracy on all datasets. When $\sigma$ further increased to 0.01, the performance decreased more noticeably, with strict and lenient F1 scores dropping by approximately 5% relative to the no-noise setting.

Overall, these results suggest that Gaussian perturbation on LoRA updates introduces a utility-privacy trade-off, i.e., smaller noise levels have little impact on model performance, while larger noise levels improve privacy preservation at the cost of reduced extraction accuracy. This ablation experiment shows that adding lightweight noise to adapter parameters is a practical privacy enhancement method in federated medical LLM training.

## 4. Discussion

This study address a central bottleneck in medical LLM development: the tension between the need for multi-institutional training and the reality of privacy constraints, clinical heterogeneity, and limited computational infrastructure. While FL has been increasingly adopted in healthcare, most existing studies have focused on smaller models such as BERT-based or logistic regression architectures. Many current publications are survey papers that primarily discuss the potential and challenges of extending FL to LLMs in medicine [26,63]. A few pioneering works have explored adapting FL to vision foundation models, primarily for disease diagnosis based on imaging data [30,64]. This gap stems from both practical and technical barriers: (1) full-parameter federated training of LLMs is computationally expensive in terms of communication and memory requirements, often exceeding the resources available in clinical environments; (2) naïve aggregation of large parameter updates is unstable under severe client heterogeneity, which is particularly common in medical data; and (3) legal, infrastructural, and operational constraints continue to hinder multi-institutional data collaboration in healthcare. Indeed, bridging FL and LLMs remains a fundamental challenge even in the general domain, as extensively discussed in recent reviews [65,66]. In addition, we also included centralized training for both BERT-based and LLM-based models as a reference. Centralized training pools data from all sites to train a single model, which is then evaluated on the same test sets. This configuration provides an empirical upper bound, representing an ideal but impractical setting in healthcare due to privacy and governance constraints. The implementation and hyperparameter details are provided in Section 5 Methods.

Our results show that federated, parameter-efficient adaptation can bridge this gap. The proposed Fed-MedLoRA and Fed-MedLoRA+ transmit only low-rank adapter updates, dramatically reducing communication costs, while Fed-MedLoRA+ further introduces adaptive, data-aware aggregation to mitigate clinical data heterogeneity. As a pilot study, this work demonstrates several key findings regarding the feasibility of federated LLMs in medicine. First, federated fine-tuning of LLMs is effective. Systematic evaluations across three settings consistently showed that both Fed-MedLoRA and Fed-MedLoRA+ outperformed the non-federated baselines, including zero-shot LLMs, single-site fine-tuned LLMs, and Bio_ClinicalBERT. In addition, both methods improve over the general-domain FL baseline FedSA-LoRA, while Fed-MedLoRA+ achieved performance comparable to centralized training. We also clarified that the JSON-format ablation study shows that these performance gains are not driven by output-template compliance.

In the new-site case study, Fed-MedLoRA+ had the best performance with a strict F1-score of 73% and a lenient F1-score of 85%, demonstrating that a new clinical site could directly benefit from federated LLMs to bootstrap local model development even with limited annotations. Federated adaption captures complementary cross-site information, leading to improved out-of-domain generalization. The results also suggest that federated LLMs can support multi-task training, where each site contributes different available task annotations. Second, the proposed approach improves both computational and communication efficiency. By transmitting only low-rank adapter parameters, communication costs were reduced by 98.5% compared with full-model fine-tuning. In practice, training federated LLMs at the 8B parameter scale using Fed-MedLoRA or Fed-MedLoRA+ requires only a single consumer-grade GPU (e.g., NVIDIA RTX 4090, 16 GB). Inference can be performed on commonly available GPUs such as the RTX A6000 (12–16 GB), RTX 3090/4090, or other high-memory workstation GPUs (≥8 GB VRAM). This efficiency improves accessibility and broadens the potential for federated LLM development in typical clinical research centers. Additional analysis on federated LLMs with the 1B backbone further showed that they can be trained with widely accessible GPUs such as the NVIDIA RTX 3060 Ti or RTX 4060 Ti (8 GB), and inference can be performed on a standard laptop (e.g., devices equipped with an Apple M3 Pro–class chip or an equivalent x86 GPU with over 18 GB RAM), with a trade-off in accuracy (for clinical IE, as the results suggested, the accuracy decrease is moderate: 3% on NER and up to 7% on RE for clinical IE)). Third, the proposed framework demonstrates the potential of scalability and robustness. Fed-MedLoRA+ maintained stable performance as the number of simulated sites increased to 10, with F1-score differences within

~2% of centralized training. Also, as previously discussed, the performance remained robust when participating sites provided different task annotations.

We evaluated the accuracy and practical feasibility of the proposed framework using clinical IE as a downstream case study. Many studies have shown that LLMs have suboptimal performance for clinical IE, often producing inconsistent predictions, incomplete extractions, and hallucinated entities or relations [16,67]. This limitation likely reflects several characteristics of clinical IE, e.g., exact span detection, abbreviation disambiguation, and structured output requirements over long and heterogeneous notes. These settings are highly sensitive to context window constraints, truncation and vulnerability to output formats [68,69]. We therefore selected clinical IE as a representative application to evaluate whether the proposed framework can improve LLM performance. In addition, IE provides a controlled benchmark for systematically assessing federated optimization, parameter-efficient adaptation, and robustness to cross-site heterogeneity. Unlike traditional encoder-based sequence-labeling settings [70,71], clinical IE may also benefit from LLMs' ability to handle heterogeneous narratives, open-vocabulary terminology, and more complex relational patterns across institutions.

To date, fine-tuning domain-specific BERT variants remains the state-of-the-art approach for clinical IE. Recent efforts have explored LLM instruction-tuning for this purpose [16,72], but most have relied on single-institution datasets or combined multiple datasets directly—approaches that overlook the privacy and governance constraints central to real-world clinical data sharing. The results demonstrate that both Fed-MedLoRA and Fed-MedLoRA+ consistently outperformed existing methods. Specifically, Fed-MedLoRA+ improved zero-shot LLM performance by up to 65% absolute F1 (e.g., from 0.345 to 0.850 for NER and from 0.203 to 0.860 for RE in two-site Llama-3.1-8B-Instruct experiments) and by ~25% over single-site fine-tuned LLMs (e.g., from 0.589 to 0.831 on NER with DeepSeek-R1-Distill-Llama-8B). Compared with fine-tuned BERT baselines, Fed-MedLoRA+ also achieved large gains—improving RE by ~70% on UTP (0.793 vs. 0.270) and ~50% on i2b2 (0.740 vs. 0.302) in the two-site experiment, with comparable improvements (~50%) in the three-site experiment (0.713 vs. 0.219). Collectively, these results demonstrate that federated learning could overcome key limitations of LLMs in clinical IE, achieving robust accuracy and cross-site generalization while maintaining data privacy.

While the proposed methods consistently outperformed BERT-based approaches across all settings in this study, it is important to recognize that BERT-based models remain more lightweight and deployment-friendly, with much fewer parameters and lower computational demands [73–75]. Achieving superior accuracy with LLMs therefore involves trade-offs that downstream users should consider when determining when to LLMs in practice. Our head-to-head comparisons highlight three scenarios where LLMs offer distinct advantages to make their best use. First, independent or heterogeneous patient cohorts. When training and testing share the same distribution, the performance gap between BERT and LLMs is modest (e.g., under the training/testing setting, Fed-MedLoRA+ was ~4% higher F1 than Bio_ClinicalBERT for NER: 0.850 vs. 0.807 on MIMIC-III; 0.866 vs. 0.827 for UTP). In contrast, on external populations, the advantage of LLMs becomes more distinct: Fed-MedLoRA+ achieved ~14.8% higher F1 on UTP (0.840 vs. 0.692) and ~9.2% higher on I2B2 (0.860 vs. 0.768). In the new-site case study, Fed-MedLoRA+ outperformed the centralized Bio_ClinicalBERT by ~13% (0.732 vs. 0.600). Second, extractive tasks requiring beyond NER. The results also demonstrate that more challenging tasks such as RE saw a consistent performance drop compared to NER. As mentioned, RE requires modeling cross-entity semantic relations over longer context. The results show that BERT-based models suffered a notable performance drop on RE, whereas LLMs had more robust performance. For instance, in the two-site experiment, Fed-MedLoRA+ achieved 0.860 F1 compared with 0.434 for Bio_ClinicalBERT. Possible explanations are twofold. First, LLMs are pretrained on much larger and more diverse corpora, whereas domain-specific BERT models are pretrained on specific medical corpora (e.g., Bio_ClinicalBERT was pretrained on MIMIC-III), which might constrain their generalization ability. Second, BERT-based models may rely more heavily on lexical and annotation patterns specific to their training datasets (e.g., abbreviations, formatting, or tagging conventions), leading to reduced robustness when these patterns differ in external datasets. Third, multi-task uses with partial task annotations. Instruction-tuned LLMs naturally support a single model capable of performing multiple tasks (e.g., both NER and RE in this study), and prior research has shown that a single LLM can adapt to a wide range of medical tasks, from extractive to generative, through instruction tuning [76,77]. Importantly, beyond their multi-task flexibility, our results demonstrate that Fed-MedLoRA and Fed-MedLoRA+ maintained robust performance even when participating sites contributed different task annotations—for example, Site A provided NER and RE labels, while Site B included only NER. Compared with their fully annotated counterparts, Fed-MedLoRA+ had only a slight decrease in NER performance (1.2%–1.7% reduction in F1-score across test sets). For RE, the performance drops were somewhat larger but still moderate, with absolute F1-score decreases of 2.8%–5.8%. This configuration mirrors realistic clinical settings, where annotation availability varies across institutions, and it is often impractical for all sites to provide complete task labels. In contrast, standard BERT-based pipelines are typically task-specific; for instance, in this study, we followed established fine-tuning protocols to train separate BERT models for NER and RE. Although multi-task BERT variants have been proposed [78], they require customized architectures and are not readily extensible to generative tasks.

Several questions remain and we highlight key aspects not addressed in the current work. First, although the proposed framework was systematically evaluated across multiple open-weight, decoder-only LLM backbones, including Llama-3.1-8B-Instruct, Llama-3.2-1B, and DeepSeek-R1-Distill-Llama-8B, under diverse accuracy settings (e.g., single-site, multi-site, and new-site adaptation) as well as feasibility settings (e.g., scalability, efficiency, and privacy-related trade-offs), its broader generalizability across the rapidly evolving landscape of LLM architectures remains to be fully established. Our experiments already provide evidence that Fed-MedLoRA and Fed-MedLoRA+ are effective across different backbone families in practical settings. However, the extent to which these findings extend to newer architectures with substantially different context lengths, parameter budgets, or generation objectives remains an open question. Broader evaluation across additional model families will help clarify the settings in which parameter-efficient federated adaptation is most effective.

Second, although our framework transmits only low-rank adapter parameters rather than full model weights or raw data, federated systems may still be vulnerable to gradient or adapter inversion [79], inadvertent memorization of sensitive data [80], and membership inference [81]. We additionally evaluated Gaussian perturbation on transmitted LoRA updates and observed a utility-privacy trade-off. Smaller noise levels have little impact on model performance, while larger noise levels reduced extraction accuracy more substantially. These findings suggest that lightweight noise injection can serve as a practical privacy-enhancing mechanism. Therefore, systematic evaluation of stronger privacy-preserving techniques (such as secure aggregation [82], client-level differential privacy [83,84], and strict access control [85]) and their efficiency trade-offs is an important direction for future medical federated LLM training and deployment.

Third, we used clinical IE as a representative downstream task to demonstrate the effectiveness of the proposed framework. However, LLMs can be adapted to a wide range of clinical tasks [7,8], and the advantage of federated LLM training may become even more significant in more open scenarios such as summarizing, question answering, and reasoning-intensive applications. Systematic evaluation across diverse medical applications and identification of which tasks and data regimes benefit most from federated learning will be critical for guiding future model development and prioritizing high-impact clinical use cases. Fourth, while this study evaluated feasibility in terms of robustness, scalability, and efficiency, real-world deployment introduces additional challenges, such as network reliability, coordination and orchestration across institutions, hardware heterogeneity, and site-specific data governance and compliance requirements [86,87]. Addressing these factors through prospective deployments and longitudinal evaluations will be essential for translating federated LLM frameworks into operational healthcare systems [88].

# 5. Methods

In this section, we first describe how LLMs are fine-tuned at a single site, followed by a detailed explanation of the proposed Fed-MedLoRA and Fed-MedLoRA+ frameworks. This case study is approved by the Yale Institutional Review Board (IRB) under protocol number 2000036010.

### 5.1 Supervised fine-tuning of LLMs in the single-site setting

The single-site setting is arguably the most common scenario in medical applications, where each site independently performs supervised fine-tuning of LLMs on its own local dataset [34,89]. In practice, data and models are not shared across sites due to the sensitivity of clinical data. Formally, client $k \in [K]$ holds a local dataset $D_k = \{(x_{k,i}, y_{k,i})\}_{i=1}^{n_k}$, where $x_{k,i}$ denotes a dataset instance and $n_k$ is the dataset size. Site $k$ will fine-tune a LLM backbone denoted as $W^{(0)} \in \mathbb{R}^{d \times l}$, where $d$ and $l$ denote the matrix dimensions.

Supervised fine-tuning of LLMs is typically performed through instruction tuning, where each dataset instance is reformulated as a unified instruction containing a task description (e.g., NER), inputs, and desired outputs with ground-truth labels (e.g., manually annotated entities) [90]. The loss function for fine-tuning all parameters of the LLM is:

$$L_k(W) = \frac{1}{n_k} \sum_{(x,y) \in D_k} l(x_{k,i}, y_{k,i}; W),$$

where $l$ denotes the cross-entropy loss between the predicted token probability distribution and the ground-truth output tokens. Because fine-tuning all parameters of an LLM is computationally expensive, in practice, parameter-efficient instruction tuning methods such as Low-Rank Adaptation (LoRA) are adopted [91]. The main idea is to only fine-tune a subset of parameters while preserving the accuracy. In LoRA, the trainable weight update $\Delta W \in \mathbb{R}^{d \times l}$ is decomposed into two low-rank matrices, $\Delta W = BA$ where $B \in \mathbb{R}^{d\times r}, A \in \mathbb{R}^{r\times l}, r \ll \min(d,l)$. During fine-tuning, only the low-rank matrices $B$ and $A$ are updated, while the original weight matrix $W$ remains frozen. The updated loss function becomes:

$$L_k(W + \Delta W) = \frac{1}{n_k}\sum_{(x,y)\in D_k} l(x_{k,i}, y_{k,i}; +\Delta W),$$

### 5.2 Proposed Fed-MedLoRA and Fed-MedLoRA+

In contrast to the single-site setting, federated learning enables collaborative model training across multiple institutions while keeping raw data local. In each round, site $k$ fine-tunes the model on its local dataset, transmits updates to a coordinating server, and receives aggregated parameters, which are then used to continue local training. We proposed two federated learning variants, Fed-MedLoRA and Fed-MedLoRA+, detailed below.

We first introduce Fed-MedLoRA, a simple yet effective algorithm inspired by the classical FedAvg [28,92]. FedAvg, which has been widely used in federated learning, aggregates full model weights from participating clients and averages them at the server [34,93]. However, for LLMs, transmitting the entire set of model weights at each round introduces significant communication overhead. To address this, Fed-MedLoRA requires clients to send only their LoRA updates. The procedure, summarized in Algorithm 1, consists of three phases.

First, the server and clients store a frozen LLM backbone (e.g., Llama-3.1-8B-Instruct) with parameters $W^{(0)} \in \mathbb{R}^{d\times l}$. The server also initializes the global LoRA modules $B^{(0)} \in \mathbb{R}^{d\times r}$ and $A^{(0)} \in \mathbb{R}^{r\times l}$ with rank $r \ll \min(d,l)$, and prepares a unified instruction format for fine-tuning. In this study, we designed instruction formats for NER and RE tasks consistent with prior work [16] (see Supplementary S1.1 and S1.2) and distributed them to all clients.

At communication round $t$, the server samples a subset $M_t$ of $m$ clients and transmits the global LoRA modules $B^{(t)}$, $A^{(t)}$ to them. Each selected client $k \in M_t$ initializes its local LoRA modules with the received values and performs $E$ iterations of local training on $D_k$, producing updated modules $B_k^{(E)} \in \mathbb{R}^{d\times r}$ and $A_k^{(E)} \in \mathbb{R}^{r\times l}$. The client then returns these updated modules to the server.

The server updates the received LoRA modules by $B_k^{(t+1)} = B_k^{(E)}$, $A_k^{(t+1)} = A_k^{(E)}$ and aggregates them to obtain the new global modules with:

$$B^{(t+1)} = \frac{1}{m}\sum_{k\in M_t}\frac{n_k}{N}B_k^{(t+1)}, A^{(t+1)} = \frac{1}{m}\sum_{k\in M_t}\frac{n_k}{N}A_k^{(t+1)},$$

where $n_k$ is the local dataset size and $N = \sum_{k=1}^{K} n_k$ is the size of all local datasets. These steps are repeated for $T$ communication rounds. The final global model is obtained by merging the frozen backbone with the aggregated LoRA modules $W^{(T)} = W^{(0)} + B^{(T)}A^{(T)}$.

Compared to transmitting and aggregating full model weights, which requires transmitting $d \times l$ parameters, Fed-MedLoRA reduces each round of communication to $dr + rl$ parameters per client. When $r \ll \min(d,l)$, this leads to substantial reductions in both communication cost and GPU memory footprint.

While Fed-MedLoRA substantially reduces the communication cost of fine-tuning LLMs in federated settings, it suffers a core limitation of FedAvg: all client updates are treated equally and aggregated via simple averaging. In clinical FL, this assumption is often violated—local datasets can vary significantly in population characteristics, class balance, annotation quality, and the prevalence of rare labels [34,89,94]. This heterogeneity (e.g., skewed label distributions, noisy or inconsistent annotations, and unrepresentative local samples) can cause simple averaging to degrade global model accuracy and amplify the influence of harmful updates [93,95].

To address this challenge, we propose Fed-MedLoRA+ (an overview is provided in Figure 1A), which dynamically estimates the influence of each client based on validation performance and performs adaptive, data-aware aggregation. Specifically, we first evaluate client updates on a small holdout validation set and compute influence scores that reflect how much each update improves or harms validation performance. These scores are then used to scale client updates during the server aggregation step. By down-weighting noisy or non-representative updates and up-weighting updates that significantly improve generalization, Fed-MedLoRA+ reduces the negative impact of local heterogeneity and builds a more robust global model on unseen sites. The procedure is summarized in Algorithm 2. As with Fed-MedLoRA, the process consists of three stages—initialization, local updates, and server aggregation—with the main distinction being the calculation of client influence and dynamic aggregation, described below.

We assume the server retains a small validation set (e.g., five clinical notes) $D_v = \{(x_{v,i}, y_{v,i})\}_{i=1}^{n_v}$ with $n_v \ll n_k$ for all clients $k \in [K]$. This assumption is standard in many federated learning deployments [93,95] and is feasible in clinical contexts, where a small amount of de-identified or publicly available labeled data can be used for validation in practice.

During the local update stage, let $M_t$ denote the subset of $m$ clients selected at round $t$, each of which sends its updated LoRA modules $B_k^{(t+1)}$ and $A_k^{(t+1)}$ to the server. Instead of averaging, the server calculates an influence score for each client. Specifically, the server first forms the client-specific model $W_k^{(t+1)} = W^{(0)} + B_k^{(t+1)} A_k^{(t+1)}$, and evaluates it on the validation set $D_v$:

$$l_k^v = -\frac{1}{n_v} \sum_{j=1}^{n_v} y_{v,j} \log P(y_{v,j} | x_{v,j}; W_k^{(t+1)})$$

where $P(\cdot)$ denotes the predictive distribution. A softmax is then applied to normalize the validation losses:

$$I_k^{(t)} = \frac{\exp(-l_k^v)}{\sum_{i \in M^{(t)}} \exp(-l_k^v)},$$

This assigns larger $I_k^{(t)}$ to clients with smaller validation loss. For numerical stability, one may compute $\exp(-l_k^v - c)$ with $c = min_i(-l_i^v)$ in practice. Each influence score is then combined with the client's local dataset size $n_k$ to form a data-aware aggregation weight:

$$C_k^{(t)} = \frac{n_k I_k^{(t)}}{\sum_{i \in M^{(t)}} n_i I_i^{(t)}}, \text{ so that } \sum_{k \in M^{(t)}} C_k^{(t)} = 1$$

Note that when all $I_k^{(t)}$ are equal, this reduces to standard FedAvg.

Using the weights $\{C_k^{(t)}\}$, the server updates the global LoRA modules as:

$$B^{(t+1)} = \frac{1}{m} \sum_{k \in M^{(t)}} C_k^{(t)} B_k^{(t+1)}, A^{(t+1)} = \frac{1}{m} \sum_{k \in M^{(t)}} C_k^{(t)} A_k^{(t+1)},$$

and the updated global model is $W^{(t+1)} = W^{(0)} + B^{(t+1)}A^{(t+1)}$.

Because the weights $\{C_k^{(t)}\}$ form a convex probability distribution, under standard smoothness and bounded-variance assumptions, Fed-MedLoRA+ inherits the same convergence guarantees as FedAvg.

**Algorithm 1:** FedMedLoRA
Input: A pre-trained LLM $W^{(0)} \in \mathbb{R}^{d\times l}$, $k \in [K]$, $T$, $E$, $m$, $D_k = \{(x_{k,i}, y_{k,i})\}_{i=1}^{n_k}$, $b$, $\eta$
Output: The optimal LoRA modules $B^*$, $A^*$

1. // Server $S$:
2. Initialize global LoRA $B^{(0)} \in \mathbb{R}^{d\times r}$, $A^{(0)} \in \mathbb{R}^{r\times l}$
3. for each round $t \in \{0, \dots, T-1\}$ do
    a) $M^t \leftarrow$ randomly select $m$ clients
    b) For each client $k \in M_t$ in parallel do
        a) $B_k^{(t+1)}, A_k^{(t+1)} \leftarrow$ LocalUpdate($k, B^{(t)}, A^{(t)}$)
    c) // Update global LoRA modules
    d) $B^{(t+1)} = \frac{1}{m}\sum_{k\in M_t}\frac{n_k}{N}B_k^{(t+1)}$, $A^{(t+1)} = \frac{1}{m}\sum_{k\in M_t}\frac{n_k}{N}A_k^{(t+1)}$
4. // Client $k \in [K]$:
5. Function $\mathrm{LocalUpdate}(k, B^{(t)}, A^{(t)})$:
    a) // Client $k$ loads global LoRA modules $B_k^{(1)} = B^{(t)}, A_k^{(1)} = A^{(t)}$
    b) For each local epoch $e \in \{1, 2, \dots, E\}$ do
        i. For each mini-batch $\gamma$ in $D_k$ of size $b$ do
            1. $W_k^{(t+1)} = W^0 + B_k^{(e)}A_k^e$
            2. $L_k^e = \frac{1}{b}\sum_{i\in\gamma} l(x_{k,i}; y_{k,i}; W_k^{(t+1)})$
            3. // Update local LoRA modules
            4. $B_k^{(e+1)} \leftarrow B_k^{(e)} - \eta\nabla_B L_k^{(e)}, A_k^{(e+1)} \leftarrow A_k^{(e)} - \eta\nabla_A L_k^{(e)}$
    c) Return $B_k^{(E)}, A_k^{(E)}$

**Algorithm 2:** FedMedLoRA+
Input: A pre-trained LLM $W^{(0)} \in \mathbb{R}^{d\times l}$, $k \in [K]$, $T$, $E$, $m$, $D_k = \{(x_{k,i}, y_{k,i})\}_{i=1}^{n_k}$, $D_v = \{(x_{k=v,i}, y_{v,i})\}_{i=1}^{v}$, $b$, $\eta$
Output: The optimal LoRA modules $B^*$, $A^*$

1. // Server $S$:
2. Initialize global LoRA $B^{(0)} \in \mathbb{R}^{d\times r}$, $A^{(0)} \in \mathbb{R}^{r\times l}$
3. for each round $t \in \{0, \dots, T-1\}$ do
    a) $M^t \leftarrow$ randomly select $m$ clients
    b) For each client $k \in M_t$ in parallel do
        i. $B_k^{(t+1)}, A_k^{(t+1)} \leftarrow$ LocalUpdate($k, B^{(t)}, A^{(t)}$)
        ii. $W_k^{(t+1)} = W^{(0)} + B_k^{(t+1)}A_k^{(t+1)}$
        iii. // Influence estimation
        iv. $l_{k,v} = -\frac{1}{n_v}\sum_{j\in D_v} y_{v,j}\log P(y_{v,j}|x_{v,j}; W_k^{(t+1)})$

v. $I_k^{(t)} = \frac{\exp(-l_{k,v})}{\sum_{k \in M_t} \exp(-l_{k,v})}$

c) // Update global LoRA modules

d) $C_k^{(t)} = \frac{n_k I_k^t}{\sum_{i \in S_t} n_i I_i^t}, B^{(t+1)} = \frac{1}{m} \sum_{k \in M_t} \frac{n_k}{N} B_k^{(t+1)}, A^{(t+1)} = \frac{1}{m} \sum_{k \in M_t} \frac{n_k}{N} A_k^{(t+1)}$

4. // Client $k \in [K]$:
5. Function LocalUpdate($k, B^{(t)}, A^{(t)}$):

d) // Client $k$ loads global LoRA modules $B_k^{(1)} = B^{(t)}, A_k^{(1)} = A^{(t)}$

e) For each local epoch $e \in \{1, 2, \dots, E\}$ do

i. For each mini-batch $\gamma$ in $D_k$ of size $b$ do

1. $W_k^{(t+1)} = W^0 + B_k^{(e)} A_k^e$
2. $L_k^e = \frac{1}{b} \sum_{i \in \gamma} l(x_{k,i}; y_{k,i}; W_k^{(t+1)})$
3. // Update local LoRA modules
4. $B_k^{(e+1)} \leftarrow B_k^{(e)} - \eta \nabla_B L_k^{(e)}, A_k^{(e+1)} \leftarrow A_k^{(e)} - \eta \nabla_A L_k^{(e)}$

f) Return $B_k^{(E)}, A_k^{(E)}$

The proposed Fed-MedLoRA and Fed-MedLoRA+ frameworks are backbone-agnostic. Two representative open-weight LLMs, Llama-3.1-8B-Instruct and DeepSeek-R1-Distill-Llama-8B, were selected as backbone models. As described in Section 4.1 Evaluation overview, we conducted head-to-head comparisons under both zero-shot and single-site fine-tuning settings using the same LLM backbones as baselines.

For single-site fine-tuning of LLMs, we applied QLoRA (4-bit quantization with low-rank adapters) to the decoder layers. Adapters were attached to all 32 decoder layers with a rank of $r$ = 16, scaling factor $\alpha$ = 64, and dropout rate of 0.05. Training was performed with a learning rate of $2 \times 10^{-4}$, 2 epochs, batch size 4, warmup ratio 0.05, and a maximum input length of 800 tokens. During inference, we set the temperature to 0 to minimize randomness [96].

For Fed-MedLoRA and Fed-MedLoRA+, we adopted the same LLM fine-tuning hyperparameters for direct comparison, with aggregation rounds set to 2. For Fed-MedLoRA+, which requires a validation set, we used 5 records for validation.

In addition, we compared against fine-tuned domain-specific BERT models (Bio_ClinicalBERT). Fine-tuning used a learning rate of $2 \times 10^{-4}$, 10 epochs, batch size 64, and a maximum input length of 512 tokens, consistent with prior studies [16].

## 5.3. Privacy preservation design

We assume that all clients including the server are semi-honest, where the server and clients follow the training protocol faithfully but may be curious about private information from transmitted model parameters. In our framework, raw clinical data remain local, while clipped LoRA adapter updates are transmitted to the server. However, these transmitted updates may still leak information through gradient or adapter inversion, membership inference, or aggregation-based information leakage. Therefore, we propose a lightweight Gaussian perturbation mechanism on the transmitted LoRA parameters.

We propose a privacy-preserving variant of Fed-MedLoRA and Fed-MedLoRA+, which are summarized in Algorithm 3 (Supplementary S3.2). The main idea is to clip the local LoRA updates and then add independent Gaussian noise to each scalar of the transmitted adapter parameters before server aggregation. Given the clipping norm $C$ and noise scale $\sigma$, we apply Gaussian noise with a standard deviation of $\sigma C$ element-wise to the clipped LoRA matrix. This mechanism is inspired by the Gaussian mechanism and noisy aggregation literature [97,98]. The proposed method provides a practical way to study the trade-off between utility and privacy under federated medical LLM training.

In Lines 18-19 of Algorithm 3, each participating client $k$ clips the local LoRA parameters to bound their sensitivity:

$$B_k^{(E)} \leftarrow B_k^{(E)} / \max(1, \frac{||B_k^{(E)}||}{C}), A_k^{(E)} \leftarrow A_k^{(E)} / \max(1, \frac{||A_k^{(E)}||}{C}).$$

In Lines 20-21, we add independent Gaussian noise to each scalar element of the clipped LoRA matrices:

$$B_k^{(E)} + N(0, \sigma^2 C^2 I), \tilde{A}_k^{(E)} = A_k^{(E)} + N(0, \sigma^2 C^2 I).$$

This noise is sampled element-wise, i.e., each scalar is perturbed independently with standard normal noise scaled by $\sigma C$.
In Lines of 6-7, the server aggregates the noisy adapter updates from participating clients using the same weighted aggregation rules as the original Fed-MedLoRA/Fed-MedLoRA+ framework. Because the noise is added before transmission, the server will never observe the unclipped local adapter parameters.

The proposed perturbation mechanism is inspired by differential privacy (DP) and noisy aggregation [97,99], but in this work we do not claim a formal $(\epsilon, \delta)$-DP guarantee. Instead, we treat the Gaussian noise scale $\sigma$ as a privacy-strength hyperparameter and empirically evaluate its effect on model utility. A formal DP guarantee would require specifying the exact sampling procedure, clipping sensitivity, training schedule, and a privacy accountant. We leave a full $(\epsilon, \delta)$-DP analysis and attack-based privacy evaluation for future work.

## Data availability

The UTP dataset cannot be shared due to privacy concerns. The MTSamples dataset can be directly accessed on the GitHub repository https://github.com/BIDS-Xu-Lab/Kiwi-LLaMA. For the MIMIC-III and 2010 i2b2 datasets, entity offsets and texts are provided without the original text files on the GitHub repository https://github.com/BIDS-Xu-Lab/Kiwi-LLaMA. Researchers can obtain the original text files from their official websites.

## Code availability

The codes are publicly available via https://github.com/Yale-BIDS-Chen-Lab/FL_LLM_Med.

## Acknowledgment

Research reported in this publication was supported by the National Library of Medicine of the National Institutes of Health under award number R01LM014604 (A.L., H.K., X.A., H.X., and Q.C.). A total of $1,127,892, or 100%, of this research was financed with federal funds. The

content is solely the responsibility of the authors and does not necessarily represent the official views of the National Institutes of Health.

## Author Contribution

A.L. and Q.C. designed the research. A.L., Y.C., W.L., Y.Y., H.Y., H.K., W.Z., Y.Z., H.P., Y.R., X.A., Z.Q., M.H., X.L., H.Y., Y.T., L.M., H.X. and Q.C. wrote and edited the manuscript. All authors contributed to discussion and manuscript preparation.

## Competing Interests

The authors declare no competing financial or non-financial interests.

Figure 1. Study overview of federated, parameter-efficient adaptation of medical LLMs under privacy, heterogeneity, and resource constraints. A. Fed-MedLoRA+ system overview. Each client locally fine-tunes a shared pretrained backbone, uploads only low-rank adapter (LoRA) updates, and receives aggregated adapters from the server; the server further computes influence scores to perform heterogeneity-aware aggregation. B. Evaluation design across three real-world settings: in-domain training/testing, external validation on independent patient cohorts, and low-resource new-site adaptation using real clinical notes from the Yale New Haven Health System. Baselines include domain-specific BERT models, proprietary and open-weight LLMs, and recent federated learning methods developed in general domains. C. Accuracy evaluation, including representative input/output examples, strict and lenient matching metrics, and manual error analysis. D. Practical feasibility evaluation, including robustness to incomplete task annotations across sites, communication and computational efficiency, scalability to increasing numbers of participating institutions, and privacy–utility trade-off analysis.

Figure 2. Resource and communication comparison across models and federated settings. (A) Communication cost (storage size) of full-parameter fine-tuning versus Fed-MedLoRA+. For each backbone, we report per-site transmission volume (bytes sent by a single client in a communication round per round, as well as the total

transmission volume for two-site and three-site experiments. (B) Peak GPU memory consumption (GB) observed when training Llama-3.2-1B, Llama-3.1-8B-Instruct, and DeepSeek-R1-Distill-Llama-8B using Fed-MedLoRA+. The horizontal dashed line marks the 16 GB of VRAM on an NVIDIA RTX 4090 for reference, demonstrating that training an 8B backbone model is feasible on high-memory consumer GPUs. (C) Training time in GPU hours for the two-site and three-site experiments.

Figure 3. Scalability evaluation: exact micro F1-scores of federated fine-tuning (Fed-MedLoRA and Fed-MedLoRA+) for Llama-3.2-1B on named entity recognition (NER) and relation extraction (RE) under different numbers of participating sites ($k$=1, 2, 3, 4, 6, 8, 10).

Table 1. Statistics of datasets, including number of documents, sentences, 4 named entity recognition (NER) entities and 16 relation extraction (RE) modifiers across different sites of datasets.

| **Count** | **Training and testing benchmark** | | | | | | **Independent validation set** | **Case study on new annotation** |
|---|---|---|---|---|---|---|---|---|
| | Training benchmark | | | Testing benchmark | | | | |
| | MIMIC-III | MTSamples | UTP | MIMIC-III | MTSamples | UTP | I2B2 | YNHH |
| Documents | 25 | 96 | 65 | 25 | 50 | 50 | 50 | 4 |
| Sentences | 4,260 | 5,211 | 4,523 | 4,049 | 2,885 | 5,778 | 4,809 | 291 |
| **Clinical entity categories** | | | | | | | | |
| Problem | 2,240 | 4,060 | 4,236 | 2,381 | 2,200 | 3,450 | 1,454 | 242 |
| Treatment | 542 | 584 | 418 | 511 | 388 | 329 | 405 | 27 |
| Test | 2,302 | 1,529 | 2,609 | 2,542 | 828 | 2,153 | 1,309 | 175 |
| Drug | 1,047 | 747 | 710 | 1,021 | 466 | 568 | 681 | 44 |
| **Total** | 6,131 | 6,920 | 7,973 | 6,455 | 3,882 | 6,500 | 3,849 | 488 |
| **Relation types** | | | | | | | | |
| Severity | 90 | 114 | 69 | 131 | 38 | 50 | 37 | |
| Temporal | 1,303 | 304 | 1,811 | 1,955 | 146 | 1,955 | 621 | |
| Negation | 467 | 1,519 | 2,339 | 614 | 800 | 1,876 | 282 | |
| Dosage | 244 | 60 | 258 | 228 | 46 | 227 | 90 | |
| Strength | 463 | 149 | 318 | 419 | 170 | 272 | 261 | |
| Reference range | 11 | 0 | 114 | 5 | 0 | 45 | 0 | |
| Uncertain | 81 | 62 | 27 | 58 | 20 | 15 | 10 | |
| Lab value | 1,579 | 893 | 1,529 | 1,954 | 397 | 1,327 | 900 | |
| Route | 346 | 104 | 351 | 329 | 89 | 292 | 245 | / |
| Frequency | 456 | 200 | 290 | 393 | 207 | 251 | 285 | |
| Subject | 19 | 26 | 128 | 26 | 12 | 131 | 0 | |
| Form | 372 | 54 | 581 | 336 | 45 | 526 | 129 | |
| Condition | 42 | 17 | 14 | 74 | 20 | 18 | 1 | |
| Duration | 67 | 18 | 20 | 25 | 10 | 21 | 27 | |
| Body location | 709 | 954 | 1,189 | 717 | 578 | 947 | 468 | |
| Course | 162 | 218 | 140 | 183 | 108 | 114 | 93 | |
| **Total** | 6,411 | 4,692 | 9,178 | 7,447 | 2,686 | 7,707 | 3,449 | |

Table 2. Strict and lenient micro F1 scores for named entity recognition (NER) and relation extraction (RE) test sets under two-site experiments. Results are reported for three backbones (Llama-3.1-8B-Instruct, DeepSeek-R1-Distill-Llama-8B and GPT-4o) and BERT baseline (Bio_ClinicalBERT) across five training strategies: zero-shot, single-site fine-tuning (average across sites, baseline), Fed-MedLoRA, Fed-MedLoRA+, and centralized fine-tuning (upper bound). All values are micro-averaged F1 scores (strict and lenient). Centralized (upper bound) rows are shaded in gray. For each model, the best non-centralized method is shown in bold and the second-best is underlined.

| Models | Methods | NER test set | | | | RE test set | | | |
|---|---|---|---|---|---|---|---|---|---|
| | | **Strict F1-scores** | | **Lenient F1-scores** | | **Strict F1-scores** | | **Lenient F1-scores** | |
| | | MIMIC-III | MTSamples | MIMIC-III | MTSamples | MIMIC-III | MTSamples | MIMIC-III | MTSamples |
| Bio_ClinicalBERT | Centralized (Upper bound) | 0.807±0.015 | 0.827±0.009 | 0.907±0.012 | 0.905±0.008 | 0.590±0.018 | 0.584±0.006 | 0.735±0.004 | 0.657±0.019 |
| | Single site (Avg, baseline) | 0.753±0.007 | 0.792±0.011 | 0.867±0.013 | 0.888±0.005 | 0.434±0.016 | 0.474±0.003 | 0.595±0.010 | 0.549±0.017 |
| GPT-4o | Zero-shot (Baseline) | 0.556±0.009 | 0.602±0.012 | 0.815±0.005 | 0.834±0.016 | 0.124±0.003 | 0.096±0.014 | 0.289±0.007 | 0.178±0.019 |
| Llama-3.1-8B-Instruct | Centralized (Upper bound) | 0.856±0.014 | 0.870±0.008 | 0.954±0.019 | 0.958±0.002 | 0.867±0.012 | 0.831±0.007 | 0.889±0.016 | 0.913±0.001 |
| | Zero-shot (Baseline) | 0.345±0.005 | 0.437±0.010 | 0.503±0.014 | 0.644±0.006 | 0.203±0.017 | 0.056±0.009 | 0.239±0.003 | 0.074±0.018 |
| | Single site (Avg, baseline) | 0.803±0.011 | 0.843±0.004 | 0.918±0.015 | 0.926±0.007 | 0.648±0.013 | 0.721±0.019 | 0.698±0.008 | 0.798±0.002 |
| | **Fed-MedLoRA (Ours)** | 0.847±0.009 | 0.858±0.016 | 0.942±0.003 | 0.929±0.012 | 0.850±0.014 | 0.817±0.005 | 0.860±0.018 | 0.849±0.007 |
| | **Fed-MedLoRA+ (Ours)** | **0.850±0.006** | **0.866±0.013** | **0.949±0.008** | **0.937±0.015** | **0.860±0.001** | **0.826±0.010** | **0.868±0.017** | **0.856±0.004** |
| DeepSeek-R1-Distill-Llama-8B | Centralized (Upper bound) | 0.852±0.012 | 0.862±0.005 | 0.952±0.019 | 0.963±0.009 | 0.863±0.014 | 0.829±0.003 | 0.842±0.016 | 0.911±0.011 |
| | Zero-shot (Baseline) | 0.291±0.008 | 0.303±0.014 | 0.438±0.006 | 0.499±0.017 | 0.163±0.002 | 0.106±0.011 | 0.190±0.019 | 0.139±0.005 |
| | Single site (Avg, baseline) | 0.797±0.013 | 0.836±0.002 | 0.902±0.010 | 0.921±0.008 | 0.589±0.015 | 0.719±0.007 | 0.644±0.004 | 0.793±0.016 |
| | **Fed-MedLoRA (Ours)** | 0.841±0.018 | 0.858±0.006 | 0.932±0.011 | 0.936±0.014 | 0.799±0.003 | 0.767±0.009 | 0.826±0.017 | 0.892±0.012 |
| | **Fed-MedLoRA+ (Ours)** | **0.845±0.004** | **0.860±0.015** | **0.945±0.007** | **0.951±0.013** | **0.831±0.010** | **0.773±0.018** | **0.831±0.001** | **0.907±0.008** |

Table 3. Strict and lenient micro F1 scores for named entity recognition (NER) and relation extraction (RE) test sets under three-site experiments. Results are reported for two open-weight LLM backbones (Llama-3.1-8B-Instruct and DeepSeek-R1-Distill-Llama-8B) and one clinical encoder baseline (Bio_ClinicalBERT) across five training strategies: zero-shot, single-site fine-tuning (average across sites, baseline), Fed-MedLoRA, Fed-MedLoRA+, and centralized fine-tuning (upper bound). All values are micro-averaged F1 scores (strict and lenient). Centralized (upper bound) rows are shaded in gray. For each model, the best non-centralized method is shown in bold and the second-best is underlined.

| Model | Method | NER test set | | | RE test set | | |
|---|---|---|---|---|---|---|---|
| | | **Strict F1-score** | | | | | |
| | | MIMIC-III | MTSamples | UTP | MIMIC-III | MTSamples | UTP |
| Bio_Clinical BERT | Centralized (Upper bound) | 0.810±0.042 | 0.838±0.015 | 0.823±0.063 | 0.598±0.027 | 0.515±0.051 | 0.413±0.009 |
| | Sigle site (Avg, baseline) | 0.782±0.068 | 0.790±0.033 | 0.720±0.058 | 0.343±0.044 | 0.433±0.021 | 0.249±0.038 |
| GPT-4o | Zero-shot (Baseline) | 0.556±0.028 | 0.602±0.040 | 0.477±0.019 | 0.124±0.019 | 0.096±0.007 | 0.091±0.063 |
| Llama-3.1-8B-Instruct | Centralized (Upper bound) | 0.840±0.017 | 0.867±0.062 | 0.909±0.036 | 0.851±0.049 | 0.788±0.011 | 0.924±0.052 |
| | Zero-shot (Baseline) | 0.345±0.053 | 0.437±0.026 | 0.256±0.070 | 0.203±0.008 | 0.056±0.039 | 0.091±0.014 |
| | Single site (Avg, baseline) | 0.775±0.061 | 0.841±0.014 | 0.798±0.047 | 0.656±0.032 | 0.703±0.055 | 0.778±0.066 |
| | **Fed-MedLoRA (Ours)** | **0.809±0.023** | 0.861±0.069 | 0.897±0.041 | 0.744±0.018 | 0.759±0.066 | **0.916±0.029** |
| | **Fed-MedLoRA+ (Ours)** | 0.803±0.037 | **0.862±0.054** | **0.898±0.012** | **0.771±0.060** | **0.762±0.029** | 0.913±0.044 |
| DeepSeek-R1-Distill-Llama-8B | Centralized (Upper bound) | 0.830±0.045 | 0.846±0.067 | 0.907±0.020 | 0.843±0.057 | 0.785±0.004 | 0.913±0.021 |
| | Zero-shot (Baseline) | 0.291±0.031 | 0.303±0.065 | 0.251±0.050 | 0.163±0.013 | 0.106±0.024 | 0.110±0.057 |
| | Single site (Avg, baseline) | 0.778±0.038 | **0.825±0.059** | 0.806±0.016 | 0.605±0.048 | 0.725±0.002 | 0.756±0.025 |
| | **Fed-MedLoRA (Ours)** | 0.799±0.056 | 0.803±0.010 | 0.848±0.064 | 0.737±0.041 | **0.739±0.025** | 0.890±0.013 |
| | **Fed-MedLoRA+ (Ours)** | **0.802±0.022** | 0.818±0.071 | **0.879±0.005** | **0.752±0.046** | 0.736±0.034 | **0.894±0.048** |
| | | **Lenient F1-score** | | | | | |
| | | MIMIC-III | MTSamples | UTP | MIMIC-III | MTSamples | UTP |
| Bio_Clinical BERT | Centralized (Upper bound) | 0.919±0.015 | 0.912±0.008 | 0.884±0.019 | 0.742±0.012 | 0.621±0.006 | 0.571±0.017 |
| | Sigle site (Avg, baseline) | 0.848±0.003 | 0.869±0.014 | 0.837±0.011 | 0.470±0.018 | 0.500±0.005 | 0.276±0.013 |
| GPT-4o | Zero-shot (Baseline) | 0.815±0.016 | 0.834±0.010 | 0.676±0.003 | 0.289±0.020 | 0.178±0.014 | 0.213±0.005 |
| Llama-3.1-8B-Instruct | Centralized (Upper bound) | 0.925±0.007 | 0.926±0.016 | 0.913±0.009 | 0.885±0.004 | 0.855±0.020 | 0.932±0.010 |
| | Zero-shot (Baseline) | 0.503±0.002 | 0.644±0.019 | 0.513±0.008 | 0.239±0.015 | 0.074±0.006 | 0.112±0.012 |
| | Single site (Avg, baseline) | 0.876±0.014 | 0.867±0.005 | 0.878±0.017 | 0.699±0.009 | 0.780±0.011 | 0.812±0.004 |
| | **Fed-MedLoRA (Ours)** | 0.893±0.010 | 0.901±0.003 | 0.906±0.016 | 0.801±0.007 | 0.797±0.019 | 0.901±0.008 |
| | **Fed-MedLoRA+ (Ours)** | **0.900±0.018** | **0.914±0.006** | **0.936±0.013** | **0.814±0.002** | **0.816±0.015** | **0.910±0.011** |
| DeepSeek-R1-Distill-Llama-8B | Centralized (Upper bound) | 0.925±0.009 | 0.926±0.020 | 0.906±0.004 | 0.885±0.017 | 0.855±0.008 | 0.932±0.014 |
| | Zero-shot (Baseline) | 0.438±0.012 | 0.499±0.007 | 0.433±0.015 | 0.190±0.010 | 0.139±0.003 | 0.128±0.018 |
| | Single site (Avg, baseline) | 0.895±0.006 | 0.855±0.019 | 0.861±0.002 | 0.732±0.014 | 0.799±0.016 | 0.807±0.001 |
| | **Fed-MedLoRA (Ours)** | 0.865±0.011 | 0.910±0.013 | 0.920±0.005 | 0.811±0.008 | 0.821±0.017 | 0.914±0.009 |
| | **Fed-MedLoRA+ (Ours)** | **0.888±0.004** | **0.925±0.018** | **0.940±0.012** | **0.890±0.019** | **0.896±0.007** | **0.918±0.002** |

Table 4. Strict micro F1 scores for named entity recognition (NER) and relation extraction (RE) (Llama-3.1-8B-Instruct) under the three-site experiment. Methods compared: FedSA-LoRA (baseline), Fed-MedLoRA and Fed-MedLoRA+.

| Method | NER test set | | | | RE test set | | | |
|---|---|---|---|---|---|---|---|---|
| | MIMIC-III | MTSamples | UTP | I2B2 | MIMIC-III | MTSamples | UTP | I2B2 |
| FedSA-LoRA (Baseline) | 0.469 | 0.476 | 0.298 | 0.365 | 0.205 | 0.108 | 0.181 | 0.184 |
| **Fed-MedLoRA (Ours)** | **0.809** | 0.861 | 0.897 | 0.833 | 0.737 | **0.739** | 0.890 | 0.711 |
| **Fed-MedLoRA+ (Ours)** | 0.803 | **0.862** | **0.898** | **0.844** | **0.752** | 0.736 | **0.894** | **0.713** |

Table 5. Strict and lenient micro F1 scores for named entity recognition (NER) test sets under the two-site experiment using HTML and JSON output formats. Results are reported for the backbone (Llama-3.1-8B-Instruct) and BERT baseline (Bio_ClinicalBERT) across five training strategies: zero-shot, single-site fine-tuning (average across sites, baseline), Fed-MedLoRA, Fed-MedLoRA+, and centralized fine-tuning (upper bound). All values are micro-averaged F1 scores (strict and lenient). Centralized (upper bound) rows are shaded in gray. For each model, the best non-centralized method is shown in bold and the second-best is underlined.

| Methods | **HTML template** | | | | **JSON template** | | | |
|---|---|---|---|---|---|---|---|---|
| | **Strict F1-scores** | | **Lenient F1-scores** | | **Strict F1-scores** | | **Lenient F1-scores** | |
| | MIMIC-III | MTSamples | MIMIC-III | MTSamples | MIMIC-III | MTSamples | MIMIC-III | MTSamples |
| Bio_ClinicalBERT | 0.807 | 0.827 | 0.907 | 0.905 | 0.795 | 0.819 | 0.901 | 0.899 |
| | 0.753 | 0.792 | 0.867 | 0.888 | 0.747 | 0.784 | 0.859 | 0.881 |
| Centralized (Upper bound) | 0.856 | 0.870 | 0.954 | 0.958 | 0.851 | 0.862 | 0.946 | 0.954 |
| Zero-shot (Baseline) | 0.345 | 0.437 | 0.503 | 0.644 | 0.336 | 0.430 | 0.496 | 0.639 |
| Single site (Avg, baseline) | 0.803 | 0.843 | 0.918 | 0.926 | 0.794 | 0.841 | 0.909 | 0.917 |
| **Fed-MedLoRA (Ours)** | 0.847 | 0.858 | 0.942 | 0.929 | 0.836 | 0.852 | 0.937 | 0.921 |
| **Fed-MedLoRA+ (Ours)** | **0.850** | **0.866** | **0.949** | **0.937** | **0.841** | **0.862** | **0.940** | **0.932** |

Table 6. Cross-institution generalization: exact micro F1-scores for named entity recognition (NER) and relation extraction (RE) across two-site and three-site experiments. Results compare zero-shot, single-site (average over sites), federated (Fed-MedLoRA, Fed-MedLoRA+) and centralized fine-tuning for Bio_ClinicalBERT, Llama-3.1-8B-Instruct and DeepSeek-R1-Distill-Llama-8B. Two-site experiment: Site A and B own MIMIC-I II and MTSamples, respectively, and external test sets are UTP and I2B2. Three-site experiment: Site A, B and C own MIMIC-III, MTSamples and UTP, respectively and test set is I2B2. Centralized (upper bound) rows are shaded in gray. For each model, the best non-centralized method is shown in bold and the second-best is underlined.

| **Model** | **Method** | **Two-site experiment** | | | | | | | | **Three-site experiment** | | | |
|---|---|---|---|---|---|---|---|---|---|---|---|---|---|
| | | NER test set | | | | RE test set | | | | NER test set | | RE test set | |
| | | Strict | | Lenient | | Strict | | Lenient | | Strict | Lenient | Strict | Lenient |
| | | UTP | I2B2 | UTP | I2B2 | UTP | I2B2 | UTP | I2B2 | I2B2 | I2B2 | I2B2 | I2B2 |
| Bio_ClinicalBERT | Centralized (Upper bound) | 0.692 | 0.768 | 0.851 | 0.838 | 0.288 | 0.354 | 0.529 | 0.536 | 0.787 | 0.863 | 0.352 | 0.579 |
| | Single site (Avg, baseline) | 0.678 | 0.747 | 0.846 | 0.808 | 0.270 | 0.232 | 0.441 | 0.419 | 0.752 | 0.816 | 0.219 | 0.495 |
| Llama-3.1-8B-Instruct | Centralized (Upper bound) | 0.852 | 0.863 | 0.932 | 0.945 | 0.803 | 0.746 | 0.887 | 0.755 | 0.852 | 0.916 | 0.745 | 0.798 |
| | Zero-shot (Baseline) | 0.256 | 0.330 | 0.513 | 0.496 | 0.091 | 0.096 | 0.116 | 0.114 | 0.330 | 0.496 | 0.096 | 0.114 |
| | Single site (Avg, baseline) | 0.806 | 0.801 | 0.884 | 0.924 | 0.730 | 0.667 | 0.822 | 0.715 | 0.756 | 0.873 | 0.659 | 0.708 |
| | **Fed-MedLoRA (Ours)** | 0.838 | 0.850 | 0.921 | 0.930 | 0.778 | 0.693 | 0.815 | 0.726 | 0.833 | **0.896** | 0.711 | 0.748 |
| | **Fed-MedLoRA+ (Ours)** | **0.840** | **0.860** | **0.927** | **0.939** | **0.793** | **0.740** | **0.854** | **0.736** | **0.844** | 0.889 | **0.713** | **0.761** |
| DeepSeek-R1-Distill-Llama-8B | Centralized (Upper bound) | 0.844 | 0.864 | 0.913 | 0.926 | 0.797 | 0.720 | 0.930 | 0.816 | 0.822 | 0.907 | 0.730 | 0.877 |
| | Zero-shot (Baseline) | 0.251 | 0.321 | 0.433 | 0.552 | 0.110 | 0.155 | 0.128 | 0.189 | 0.321 | 0.552 | 0.155 | 0.189 |
| | Single site (Avg, baseline) | 0.788 | 0.769 | 0.892 | 0.906 | 0.687 | 0.665 | 0.719 | 0.688 | 0.789 | 0.862 | 0.649 | 0.729 |
| | **Fed-MedLoRA (Ours)** | 0.819 | 0.845 | 0.901 | 0.915 | **0.789** | 0.711 | 0.901 | 0.782 | 0.791 | 0.868 | **0.692** | **0.764** |
| | **Fed-MedLoRA+ (Ours)** | **0.825** | **0.855** | **0.912** | **0.917** | 0.751 | **0.724** | **0.928** | **0.801** | **0.799** | **0.896** | 0.685 | 0.761 |
| GPT-4o | Zero-shot | 0.477 | 0.602 | 0.676 | 0.766 | 0.091 | 0.106 | 0.213 | 0.192 | 0.602 | 0.766 | 0.106 | 0.192 |

Table 7. Evaluation on Yale New Haven Health (YNHH) clinical notes for named entity recognition (NER). The table reports strict and lenient micro F1 scores, precision and recall across two-site and three-site experiments. Results compare a BERT baseline (Bio_ClinicalBERT) and the Llama-3.1-8B-Instruct backbone under four strategies: zero-shot, Fed-MedLoRA, Fed-MedLoRA+ and centralized fine-tuning (upper bound). Centralized (upper bound) rows are shaded in gray. For each model, the best non-centralized method is shown in bold and the second-best is underlined.

| **Model** | **Method** | **Metrics** | | | | | |
|---|---|---|---|---|---|---|---|
| | | Strict | | | Lenient | | |
| | | F1 score | Precision | Recall | F1 score | Precision | Recall |
| **Two-site experiment** | | | | | | | |
| Bio_ClinicalBERT | Centralized (Upper bound) | 0.596 | 0.624 | 0.571 | 0.645 | 0.676 | 0.617 |
| Llama-3.1-8B-Instruct | Centralized (Upper bound) | 0.732 | 0.778 | 0.690 | 0.823 | 0.881 | 0.772 |
| | Zero-shot (Baseline) | 0.397 | 0.348 | 0.462 | 0.524 | 0.459 | 0.609 |
| | **Fed-MedLoRA (Ours)** | <u>0.708</u> | <u>0.759</u> | <u>0.664</u> | <u>0.794</u> | <u>0.850</u> | <u>0.745</u> |
| | **Fed-MedLoRA+ (Ours)** | **0.720** | **0.764** | **0.681** | **0.809** | **0.867** | **0.758** |
| **Three-site experiment** | | | | | | | |
| Bio_ClinicalBERT | Centralized (Upper bound) | 0.606 | 0.631 | 0.583 | 0.646 | 0.669 | 0.624 |
| Llama-3.1-8B-Instruct | Centralized (Upper bound) | 0.733 | 0.755 | 0.711 | 0.858 | 0.858 | 0.859 |
| | Zero-shot (Baseline) | 0.397 | 0.348 | 0.462 | 0.524 | 0.459 | 0.609 |
| | **Fed-MedLoRA (Ours)** | <u>0.722</u> | <u>0.718</u> | <u>0.726</u> | <u>0.848</u> | <u>0.843</u> | <u>0.853</u> |
| | **Fed-MedLoRA+ (Ours)** | **0.730** | **0.732** | **0.729** | **0.854** | **0.852** | **0.856** |

Table 8. Named entity recognition (NER) error categories observed in 200 manually curated error instances. Count indicates the number of times each error category was observed across the sampled set (instances may be counted under multiple categories). Examples show ground-truth annotation (Ground-truth) and model prediction (Predict).

| **Categories** | **Count** | **Descriptions** | **Examples** |
|---|---|---|---|
| Boundary/ span error (partial match) | 90 | Predicted span overlaps the ground-truth but is shorter or longer | *Ground-truth:* <span class="test"> His urine</span> and <span class="test"> serum tox screens</span> ... *Predict:* <span class="test">His urine and serum tox screens</span> ... |
| False negative | 72 | Model fails to predict an entity that is present in the ground-truth | *Ground-truth:* He was admitted to the CMED , intubated for <span class="treatment"> airway protection</span> … *Predict:* He was admitted to the CMED , intubated for airway protection … |
| Type confusion | 34 | Correct span but wrong entity label | *Ground-truth:* 2012-06-07 09 : 55 PM <span class="test">BLOOD ASA</span> - NEG Ethanol … *Predict:* 2012-06-07 09 : 55 PM <span class="drug">BLOOD ASA</span> - NEG Ethanol … |
| False positive | 18 | Model predicts an entity where the ground-truth has none | *Ground-truth:* … above the thoracic inlet approximately 6 . 5 cm above the carina . *Predict:* … above the <span class="test">thoracic </span> inlet approximately 6 . 5 cm above the carina . |
| Boundary & type error | 16 | Wrong span and wrong entity label | *Ground-truth:* … her ST segments with <span class="test">evolution of Q waves</span> . *Predict:* … <span class="test">her ST segments</span> with evolution of <span class="problem">Q waves</span> . |
| Annotation error | 9 | The annotator annotates the wrong entity label | *Ground-truth:* Patient had No <span class="problem">Known Allergies to Drugs</span> … *Predict:* Patient had No <span class="problem">Known Allergies</span> to <span class="drug">Drugs</span> … |
| Merged/ split entities | 5 | Model merges two adjacent entities into one, or splits one ground-truth entity into multiple ones | *Ground-truth*: <span class="drug">The nasogastric tube<span> terminates within the stomach . Predict: <span class="drug">The nasogastric</span> <span class="drug">tube </span>terminates within the stomach . |

Table 9. Model robustness results on uneven task annotations from different sites: strict micro F1 scores of Fed-MedLoRA, Fed-MedLoRA+, zero-shot, single-site and centralized learning on NER and RE tasks by fine-tuning on Llama-3.1-8B-Instruct model under two-site and three-site experiments. Centralized (upper bound) rows are shaded in gray. For each model, the best non-centralized method is shown in bold and the second-best is underlined.

| Training set | Methods | NER test set | | | | RE test set | | | |
|---|---|---|---|---|---|---|---|---|---|
| | | MIMIC-III | MTSamples | UTP | I2B2 | MIMIC-III | MTSamples | UTP | I2B2 |
| **Two-site experiment** | | | | | | | | | |
| A: MIMIC-III (NER+RE)<br>B: MTSamples (NER+RE) | Centralized (Upper bound) | 0.856 | 0.870 | 0.852 | 0.863 | 0.867 | 0.831 | 0.803 | 0.746 |
| | Zero-shot (Baseline) | 0.345 | 0.437 | 0.256 | 0.330 | 0.203 | 0.056 | 0.091 | 0.096 |
| | Single site (Avg, baseline) | 0.803 | 0.843 | 0.806 | 0.801 | 0.648 | 0.721 | 0.730 | 0.667 |
| | Fed-MedLoRA (Ours) | 0.847 | 0.858 | 0.838 | 0.850 | 0.850 | 0.817 | 0.778 | 0.693 |
| | Fed-MedLoRA+ (Ours) | **0.850** | **0.866** | **0.840** | **0.860** | **0.860** | **0.826** | **0.793** | **0.740** |
| A: MIMIC-III (NER+RE)<br>B: MTSamples (NER) | Fed-MedLoRA (Ours) | 0.827 | 0.842 | 0.813 | 0.839 | 0.815 | 0.744 | 0.701 | 0.675 |
| | Fed-MedLoRA+ (Ours) | 0.838 | 0.851 | 0.823 | 0.846 | 0.832 | 0.768 | 0.741 | 0.709 |
| **Three-site experiment** | | | | | | | | | |
| A: MIMIC-III (NER+RE)<br>B: MTSamples (NER+RE)<br>C: UTP (NER+RE) | Centralized (Upper bound) | 0.840 | 0.867 | 0.909 | 0.852 | 0.851 | 0.788 | 0.924 | 0.745 |
| | Zero-shot (Baseline) | 0.345 | 0.437 | 0.256 | 0.330 | 0.203 | 0.056 | 0.091 | 0.096 |
| | Single site (Avg, baseline) | 0.775 | 0.841 | 0.798 | 0.756 | 0.656 | 0.703 | 0.778 | 0.659 |
| | Fed-MedLoRA (Ours) | **0.809** | 0.861 | 0.897 | 0.833 | 0.744 | 0.759 | **0.916** | 0.711 |
| | Fed-MedLoRA+ (Ours) | 0.803 | **0.862** | **0.898** | **0.844** | **0.771** | **0.762** | 0.913 | **0.713** |
| A: MIMIC-III (NER+RE)<br>B: MTSamples (NER+RE)<br>C: UTP (NER) | Fed-MedLoRA (Ours) | 0.775 | 0.837 | 0.828 | 0.784 | 0.681 | 0.714 | 0.765 | 0.683 |
| | Fed-MedLoRA+ (Ours) | 0.781 | 0.843 | 0.816 | 0.797 | 0.742 | 0.729 | 0.772 | 0.694 |

Table 10. Strict micro F1 scores of Fed-MedLoRA and Fed-MedLoRA+ on named entity recognition (NER) and relation extraction (RE) tasks by fine-tuning on Llama-3.2-1B and Llama-3.1-8B-Instruct models under two-site and three-site experiments.

| Training set | Methods | NER test set | | | | RE test set | | | |
|---|---|---|---|---|---|---|---|---|---|
| | | MIMIC-III | MTSamples | UTP | I2B2 | MIMIC-III | MTSamples | UTP | I2B2 |
| **Two-site experiment** | | | | | | | | | |
| Llama-3.2-1B | Fed-MedLoRA (Ours) | 0.814 | 0.821 | 0.802 | 0.834 | 0.815 | 0.798 | 0.757 | 0.677 |
| | Fed-MedLoRA+ (Ours) | 0.821 | 0.829 | 0.817 | 0.831 | 0.831 | 0.802 | 0.774 | 0.728 |
| Llama-3.1-8B-Instruct | Fed-MedLoRA (Ours) | 0.847 | 0.858 | 0.838 | 0.850 | 0.850 | 0.817 | 0.778 | 0.693 |
| | Fed-MedLoRA+ (Ours) | **0.850** | **0.866** | **0.840** | **0.860** | **0.860** | **0.826** | **0.793** | **0.740** |
| **Three-site experiment** | | | | | | | | | |
| Llama-3.2-1B | Fed-MedLoRA (Ours) | 0.779 | 0.819 | 0.854 | 0.819 | 0.734 | 0.735 | 0.886 | 0.704 |
| | Fed-MedLoRA+ (Ours) | 0.785 | 0.814 | 0.866 | 0.823 | 0.742 | 0.741 | 0.895 | 0.701 |
| Llama-3.1-8B-Instruct | Fed-MedLoRA (Ours) | **0.809** | 0.861 | 0.897 | 0.833 | 0.744 | 0.759 | **0.916** | 0.711 |
| | Fed-MedLoRA+ (Ours) | 0.803 | **0.862** | **0.898** | **0.844** | **0.771** | **0.762** | 0.913 | **0.713** |

Table 11. Strict and lenient micro F1 scores for named entity recognition (NER) using the Llama-3.1-8B-Instruct backbone under the two-site experiment with different Gaussian noise levels in privacy-preserved Fed-MedLoRA.

| Noise level | Strict F1-score | | | | Lenient F1-score | | | |
|---|---|---|---|---|---|---|---|---|
| | MIMIC-III | MTSamples | UTP | I2B2 | MIMIC-III | MTSamples | UTP | I2B2 |
| $\sigma = 0$ | 0.847 | 0.858 | 0.838 | 0.850 | 0.942 | 0.929 | 0.921 | 0.930 |
| $\sigma = 0.001$ | 0.845 | 0.857 | 0.834 | 0.848 | 0.940 | 0.927 | 0.917 | 0.926 |
| $\sigma = 0.005$ | 0.842 | 0.853 | 0.827 | 0.839 | 0.931 | 0.918 | 0.904 | 0.921 |
| $\sigma = 0.01$ | 0.789 | 0.802 | 0.793 | 0.788 | 0.904 | 0.876 | 0.865 | 0.894 |